\documentclass[journal]{IEEEtran}
\usepackage{amsmath}
\usepackage{amssymb}
\usepackage{multirow}
\usepackage{graphicx}
\usepackage{subfigure}
\usepackage{algorithm}
\usepackage{algpseudocode}
\usepackage{xcolor}
\usepackage{balance}
\usepackage{fancybox}
\usepackage{hyperref}
\usepackage{cleveref}
\newcommand{\rmk}[1]{{\color{black}#1}}

\let\olditem\item\renewcommand{\item}[1][black]{\color{#1}\olditem}

\begin{document}
\title{AAKT: Enhancing Knowledge Tracing with\\Alternate Autoregressive Modeling} 
\author{Hao Zhou, Wenge Rong, Jianfei Zhang, Qing Sun, Yuanxin Ouyang, and Zhang Xiong

\thanks{Manuscript received xx xxxx xxxx; revised xx xxxx xxxx, xx xxxx xxxx and xx xxx xxxx; accepted xx xxxx xxxx. Date of publication xx xxxx xxxx; date of current version xx xxxx xxxx. This work was supported in part by the National Natural Science Foundation of China under Grant 62477001. % and in part by the State Key Laboratory of Complex and Critical Software Environment under Grant CCSE-2024ZX-16. 
(Corresponding author: Wenge Rong.)}

\thanks{H. Zhou, W. Rong, J. Zhang, Q. Sun, and Y. Ouyang are with the School of Computer Science and Engineering, Beihang University, Beijing 100191, China. (e-mail: h.zhou@buaa.edu.cn, w.rong@buaa.edu.cn, zhangjf@buaa.edu.cn, sunqing@buaa.edu.cn, oyyx@buaa.edu.cn).

Z. Xiong is with the School of Information Technology \& Management, University of International Business and Economics, Beijing
100029, China. e-mail: xiongz@uibe.edu.cn.

Digital Object Identifier 10.1109/TLT.xxxx.xxxxxxx
}% <-this % stops a space
}

\markboth{IEEE Transactions on Learning Technologies,~Vol.~xx, No.~x, xxxx~xxxx}%
{Shell \MakeLowercase{\textit{et al.}}: Bare Demo of IEEEtran.cls for IEEE Journals}

\maketitle
\begin{abstract}
Knowledge Tracing (KT) aims to predict students' future performances based on their former exercises and additional information in educational settings. KT has received significant attention since it facilitates personalized experiences in educational situations. Simultaneously, the autoregressive modeling on the sequence of former exercises has been proven effective for this task. One of the primary challenges in autoregressive modeling for Knowledge Tracing is effectively representing the anterior (pre-response) and posterior (post-response) states of learners across exercises. Existing methods often employ complex model architectures to update learner states using question and response records. In this study, we propose a novel perspective on knowledge tracing task by treating it as a generative process, consistent with the principles of autoregressive models. We demonstrate that knowledge states can be directly represented through autoregressive encodings on a question-response alternate sequence, where model generate the most probable representation in hidden state space by analyzing history interactions. This approach underpins our framework, termed Alternate Autoregressive Knowledge Tracing (AAKT). Additionally, we incorporate supplementary educational information, such as question-related skills, into our framework through an auxiliary task, and include extra exercise details, like response time, as additional inputs. Our proposed framework is implemented using advanced autoregressive technologies from Natural Language Generation (NLG) for both training and prediction. Empirical evaluations on four real-world KT datasets indicate that AAKT consistently outperforms all baseline models in terms of AUC, ACC, and RMSE. Furthermore, extensive ablation studies and visualized analysis validate the effectiveness of key components in AAKT.
\end{abstract}


\IEEEpeerreviewmaketitle

% \section*{Nomenclature}
% \nomenclature{$\mathcal{Q}=\{q_1,q_2,\dots\}$ }{Question set}
% \nomenclature{$\mathcal{T}=\{t_1,t_2,\dots\}$}{Skill set}
% \nomenclature{$q_i$}{The i-th question of $\mathcal{Q}$}
% \nomenclature{$t_i$}{The i-th skill of $\mathcal{T}$}
% \nomenclature{$\mathcal{I}=\{\mathrm{Ex}_1,\mathrm{Ex}_2,\dots\}$}{The exercise sequence of a student}
% \nomenclature{$\tilde{q}_i\in\mathcal{Q}$}{The question in $\mathcal{I}$ at step $i$}
% \nomenclature{$\tilde{t}_i\subset\mathcal{T}$}{The skill \textbf{set} in $\mathcal{I}$ at step $i$}
% \nomenclature{$\tilde{a}_i\in\mathbb{R}$}{The posterior information in $\mathcal{I}$ at step $i$}
% \nomenclature{$\tilde{c_i}\in \{0,1\}$}{The correctness in $\mathcal{I}$ at step $i$}
% \nomenclature{$\mathrm{Ex}_j=(\tilde{q}_j,\tilde{t}_j,\tilde{a}_j,\tilde{c}_j)\in\mathcal{I}$ }{The element of $\mathcal{I}$ at step $j$}
% \nomenclature{$L_{max}$ }{Max sequence length}
% \nomenclature{$r_o\in[0,1]$}{Overlap ratio}
% \nomenclature{$L_{batch}$}{Batch size}

% \printnomenclature

\section{introduction}
\label{sec:intro}
\IEEEPARstart
{I}{t} is imperative for intelligent tutoring systems and online learning platforms to effectively depict the evolving knowledge status of students. To address this requirement, the fundamental task of Knowledge Tracing (KT) has been introduced \cite{BKT}. KT involves tracing the mastery level of individual students over time and generating predictions regarding the accuracy of their responses to diverse questions based on their knowledge status. This process entails collecting the historical exercise sequences of students, computing their mastery levels, and forecasting the likelihood of correct responses to subsequent exercises. By executing Knowledge Tracing, educators gain insights that facilitate the provision of personalized learning materials, identification of weaknesses, and targeted exercise recommendations. 

Conventional methodologies within this domain mainly include Bayesian Knowledge Tracing (BKT) employing Hidden Markov Models \cite{BKT, IndividualizedBKT, BKTOverview}, and Deep Knowledge Tracing (DKT) utilizing Deep Neural Networks \cite{DKT} along with its derivative models \cite{Survey1, Survey2}. While the initial Knowledge Tracing models, such as BKT, exhibit ease of implementation and enhanced interpretability, their efficacy is hindered by their inability to capture the dynamic nature of knowledge status due to oversimplified assumptions concerning the learning process. 

The initial DKT models use RNNs, particularly LSTM and GRU architectures, to represent students' knowledge status \cite{DKT, DKT+, DKVMN}. However, RNN-based models concentrate on nearby exercises, limiting the ability to model long-term dependencies. To address this limitation, Sequential Key-Value Memory Networks (SKVMN) \cite{SKVMN} incorporate LSTM with hops to capture extended dependencies within exercise sequences. Despite this improvement, it does not fully overcome the challenge posed by the limited attention span. The Transformer architecture \cite{Attention}, known for its effectiveness in sequence-to-sequence prediction tasks with the attention mechanism, influences certain KT models. These models integrate attention by incorporating a self-attention mechanism into Deep Knowledge Tracing (DKT) \cite{SAKT, AKT, DTrKT}. The key advantage of self-attention-based models lies in their ability to predict individual interactions by considering the entire historical exercise sequence. This addresses concentration imbalances and enhances the holistic perspective during predictions.

Despite advancements in DNNs, current KT models still exhibit limitations which can be attributed to three factors. \rmk{Firstly, some prior knowledge tracing models encode question sequence and history interaction sequences separately \cite{SAKT, SAINT+, AKT, VanillaTrKT}. These models initially encode the information of history interactions and the next question into hidden states separately. This late fusion method neglects shallow interactions and combines the two types of information only after they have been extracted into highly abstract states, resulting in a loss of bilateral information. Additionally, empirical findings reveal that, despite having more parameters, the self-attention-based AKT model fails to outperform the RNN-based DKT model on certain datasets \cite{AKT}. AKT still adheres to the paradigm of late fusion, where the ``Knowledge Retriever'' models history interactions individually. Furthermore, the monotonic attention mechanism leaves history responses unattended, resulting in the failure to effectively integrate information from both questions and responses.}

\rmk{Secondly, current KT models aiming to merge information pertaining to both questions and skills, exhibit a shortfall in meticulous fusion treatment. Notably, some models opt for a simplistic concatenation or addition of embedding vectors without considering hierarchical distinction between questions and skills \cite{SAINT+}, \cite{ATKT}. It is important that  questions and skills occupy disparate levels of abstraction and hierarchy within students' learning trajectories. Skills invariably hold a higher position than questions, as questions are derived from skills. Treating them as interchangeable overlooks the hierarchical disparity inherent in their relationship.}

\rmk{Lastly, current KT models fail to utilize datasets optimally.  Each element within a given exercise sequence is used only once during training \cite{DKT}, \cite{DKT+}, \cite{SAKT}. Dividing subsequences in a consecutive manner may overlook those with diverse distributions in the context of knowledge tracing. KT models learn how to effectively capture students' knowledge states via different subsequences of history interactions. Therefore, introducing overlaps within subsequences leads to enhanced model performance.} Furthermore, many previous KT models merely utilize the information of questions or skills \cite{DKT}, \cite{DKVMN}, \cite{AKT}. However, certain dataset components, including antecedent information known prior to responding and subsequent information discernible only after responding, remain untapped. These overlooked elements have utility in extracting students' knowledge statuses and predicting future interactions.

\begin{figure}[!htpb]
\centering
\includegraphics[width=\columnwidth]{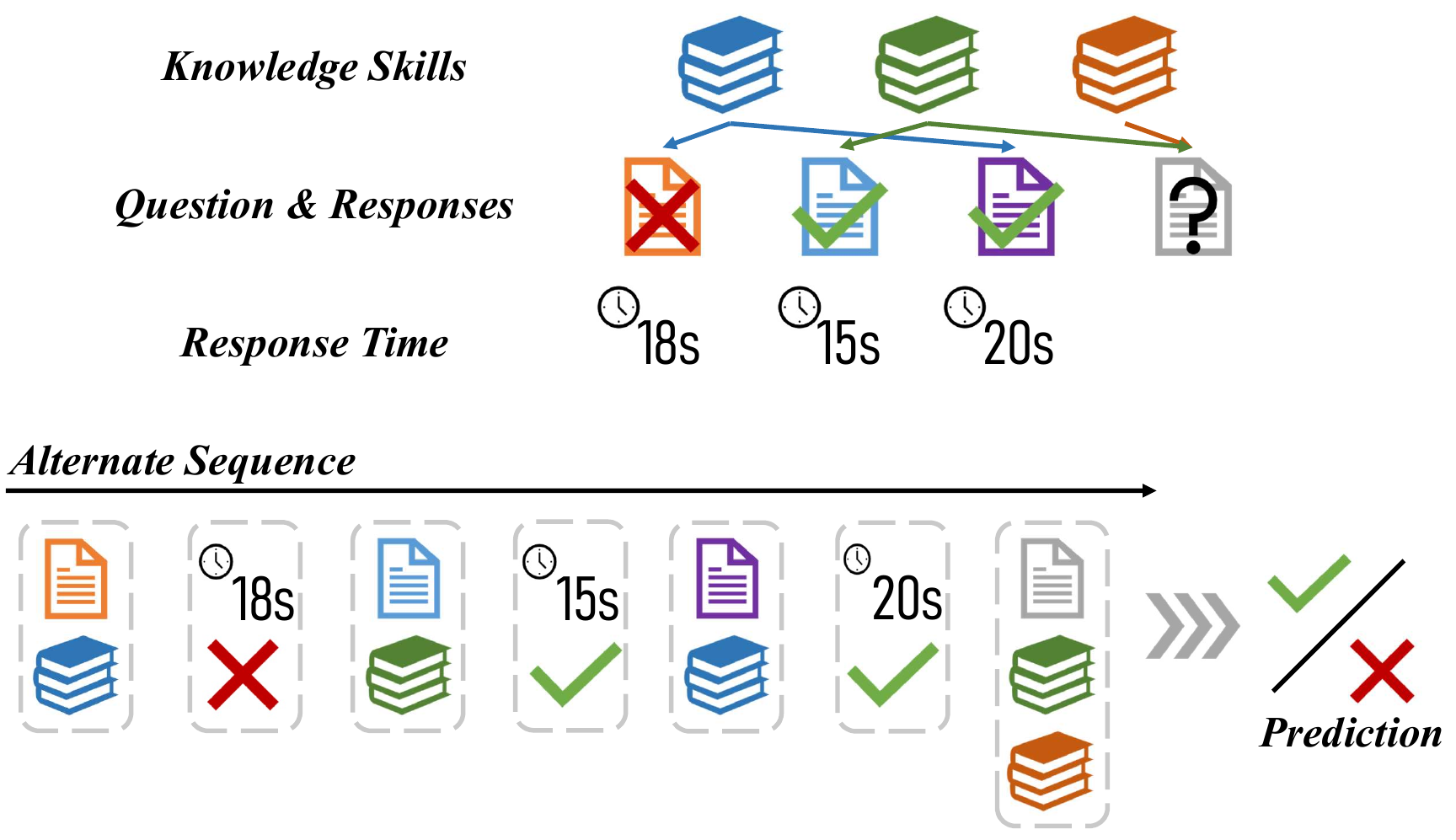}
\caption{\rmk{An overview of AAKT. Every element in students' history interactions is split into anterior information and posterior information. Subsequently, they are organized in an alternate manner, facilitating autoregressive modeling.}}
\label{fig:illustrative example}
\end{figure}

Considering the aforementioned limitations of current KT models, this paper introduces the Alternate Autoregressive Knowledge Tracing model (AAKT) as a viable solution. An illustrative example of AAKT is shown in Fig. \ref{fig:illustrative example}.

The AAKT model adopts an innovative approach by implementing alternate sequence construction, wherein elements from each question sequence and its corresponding response sequence are interleaved. This technique is augmented by the incorporation of sliding windows, enhancing dataset volume and bolstering predictive performance. Integration of information related to questions and skills is accomplished through an auxiliary task coupled with the computation of auxiliary loss. Additionally, the embedding process incorporates supplementary details, such as the time spent on answering each question, further enriching the model's capacity for comprehensive knowledge tracing.

In AAKT, a distinctive approach is employed to rearrange and combine each question sequence and its corresponding response sequence. This involves interleaving question indices and responses in an alternate order, deliberately segregating anterior and posterior information and thereby facilitating an autoregressive prediction process. The resulting rearranged sequences are partitioned into subsequences with overlaps using sliding windows. Then these subsequences are amalgamated with additional information, such as the time spent on answering questions, to generate embedding vectors. These vectors undergo an auxiliary task, ensuring their inclusion of hierarchical information pertaining to the corresponding skill(s).

The modified vectors are then fed into an autoregressive transformer, specifically the GPT-J (without pre-trained weight) instance in this paper. The output of the autoregressive transformer, representing the hidden knowledge status of students, is further transformed into prediction results via feed-forward networks. The total loss of the model is calculated as a combination of binary cross-entropy loss and Kullback-Leibler divergence generated from the auxiliary task.

The contribution of our paper can be summarized as follows:

1. \textbf{Autoregressive Perspective of Knowledge Tracing}: In this paper, we advance a novel perspective on knowledge tracing by conceptualizing it within a generative framework that aligns with the principles of autoregressive models. We posit that, within this generative paradigm, the model utilizes students' historical interactions as inputs to produce the most probable representation of the subsequent step in the hidden state space. Consequently, the inferred representation can be analytically translated into a probabilistic estimation of the correctness of the student's response to the next question. This paradigm shift provides a new understanding of the dynamic learning processes underlying student interactions.

2. \textbf{Alternate Sequence Construction}: Our work stands out as the first by introducing the innovative use of alternate sequences to integrate question sequences and response sequences. This novel methodology promotes autoregressive modeling, facilitating an automatic and effective interaction between these two essential types of information. By leveraging this approach, we enhance the model's capacity to effectively synthesize and interpret the interdependent dynamics of questions and responses.

3. \textbf{Skill-Enriched Embeddings}: We introduce skill-related information into question embeddings through an auxiliary task, thereby acknowledging the hierarchical distinction between questions and skills within the learning trajectory. This method demonstrates clear superiority over traditional embedding fusion strategies, such as addition or concatenation. By embedding skill-related data, we enhance the model's ability to capture the interplay between questions and underlying skills.

\section{Related Work}
Corbett and Anderson introduced Bayesian Knowledge Tracing (BKT) as the initial knowledge tracing model, utilizing binary variables to predict a student's mastery or non-mastery of skills within a set \cite{BKT}. BKT relies on assumptions about variable distributions, including response accuracy and question difficulty \cite{BKTAssumption1}, \cite{BKTAssumption2}. The model considers four factors (i.e., initial knowledge status, learning rate, guessing probability, and slipping probability) to predict knowledge status based on the most recent response. Subsequent research expanded BKT by introducing factors like problem difficulty or individual learning ability \cite{BKTImprovement1}, \cite{BKTImprovement2}. Additionally, Knowledge Tracing Machine (KTM) \cite{KTM} employs factorization machines \cite{FactorizationMachine} to enhance the interpretability and performance of BKT. However, the interpretability of BKT models relies on simplified assumptions, limiting their ability to capture dynamic knowledge status and provide comprehensive explanations of students' learning processes, thereby negatively impacting overall model performance.

To address the limitations of traditional KT models, Piech et al. \cite{DKT} pioneered the integration of deep neural networks with knowledge tracing, introducing the DKT model. DKT employed Long Short-Term Memory recurrent neural networks (LSTM) \cite{LSTM} to model students' knowledge status, surpassing the performance of BKT and its extensions. Seeking enhanced prediction consistency in exercise sequences, DKT+ \cite{DKT+} improved upon DKT by introducing regularization on the LSTM's hidden state, smoothing the evolution of knowledge status. Subsequent to DKT, researchers delved further into the application of neural networks in knowledge tracing. Zhang et al. \cite{DKVMN} proposed a Dynamic Key-Value Memory Network (DKVMN) employing external memory matrices to store knowledge memory and update corresponding mastery levels. Building on DKVMN, Abdelrahman and Wang \cite{SKVMN} introduced a modified LSTM, hop-LSTM, in their model. Zhu et al. \cite{StableKT} first employs causal inference for explanatory analysis of knowledge tracing, then proposes a learning algorithm for stable KT based on the analysis outcomes. Guo et al. \cite{ATKT} utilized adversarial training in Adversarial Knowledge Tracing (ATKT) to enhance model robustness and mitigate overfitting in small datasets. Wang et al. \cite{DCD} integrated educational priors and RNN in a KT model to improve interpretability.

Building on the success of attention mechanisms and transformers \cite{Attention} in natural language processing and computer vision, self-attention-based knowledge tracing models have emerged. Pandey and Karypis \cite{SAKT} were pioneers in introducing the self-attention mechanism to Knowledge Tracing models (SAKT). This innovation enables SAKT to capture long-term dependencies within historical exercises. Pu et al. \cite{VanillaTrKT} proposed a KT model based on the vanilla transformer. Additionally, Ghosh et al. \cite{AKT} presented a context-aware attentive KT model featuring monotonic attention and an exponential decay mechanism. More recently, Cui et al. \cite{MRTKT} introduced a multi-relational transformer designed to facilitate fine-grained interaction modeling between question-response pairs. Qiu et al. \cite{OPKT} proposed the optimized pretraining deep knowledge tracing (OPKT) method, which enhances knowledge state assessment by self-supervised learning, comprehensive contextual encodings and extracting latent features from learning sequences.

{Despite notable advancements, self-attention-based KT models encounter challenges in effectively integrating information from questions and responses. While various design features attempt to model both types of information, the interactive relationship between them is partly neglected. In SAKT \cite{SAKT}, history interactions, including questions and responses, and future questions are encoded into different sequences, leading to inconsistencies in semantic spaces. AKT \cite{AKT} calculates self-attention within question sequences and response sequences separately, employing a knowledge retriever to partially combine the two types of information. However, the information from responses does not contribute to calculating attention weights in knowledge retrievers, resulting in inadequate information interaction. MRT-KT \cite{MRTKT} establishes multiple relations between different exercises based on their question skill sets and response results. However, it omits including response information as encoded input due to the lack of an effective autoregressive sequence construction method.}

\begin{figure*}[ht]
\centering
\includegraphics[width=0.98\textwidth]{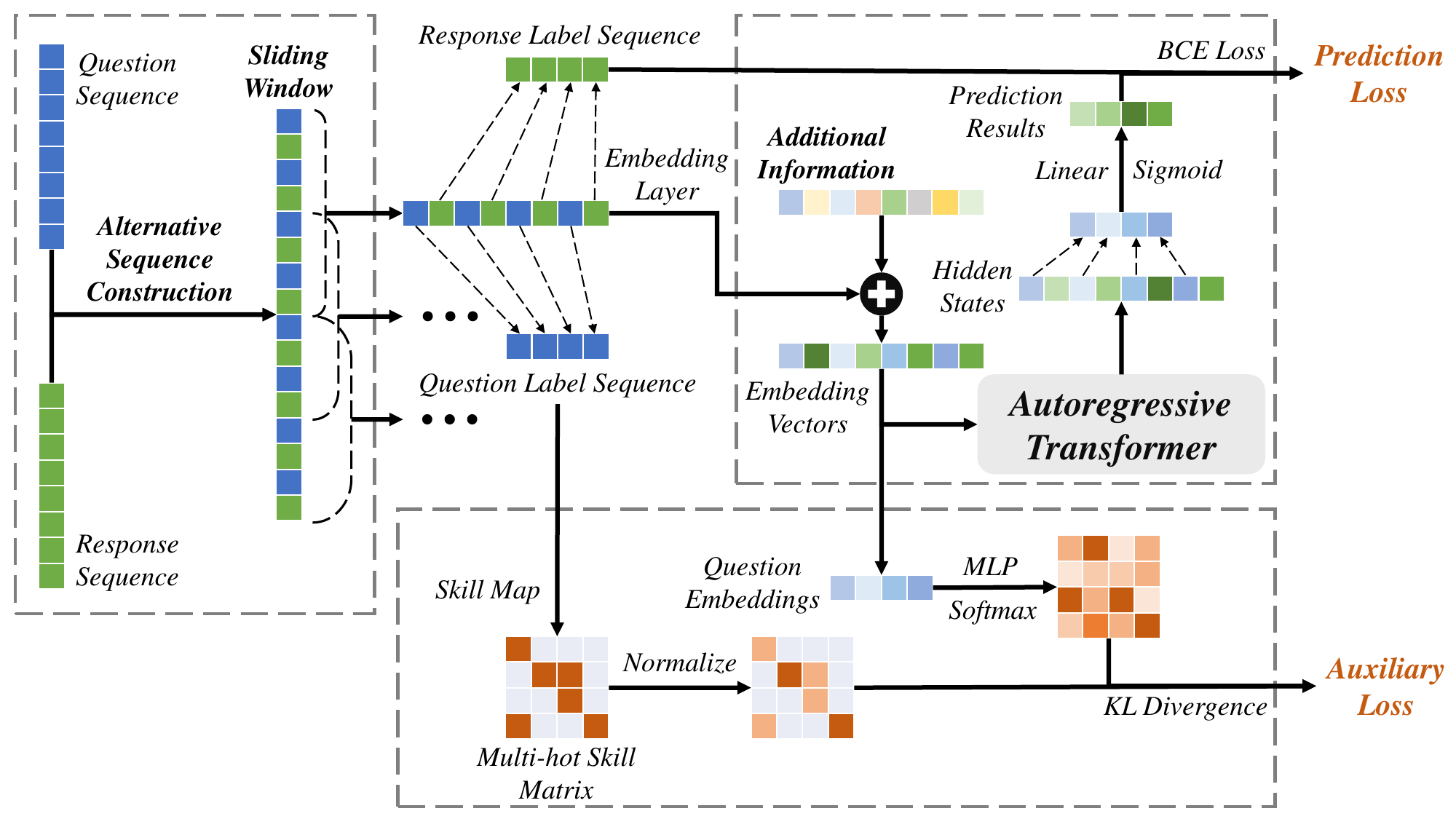}
\caption{The overall architecture of the proposed AAKT framework. It comprises three integral components. }
\label{fig:model structure}
\end{figure*}

{In response to the identified limitations of self-attention-based KT models, we advocate for better integration of questions and responses. Subsequently, we conducted an analysis of the causal relationships inherent in knowledge tracing tasks. Our proposal introduces an alternative autoregressive modeling approach based on a novel sequential representation that encompasses both question and response information.}

\section{Methodology}
{In this section, we present the details of our AAKT framework. The overall architecture is depicted in Fig. \ref{fig:model structure}. The initial segment encompasses \textbf{Alternate Sequence Construction}, a process that amalgamates the question sequence with the question and response sequences to form a distinctive sequence. This is succeeded by the \textbf{Sliding Window} mechanism, which selects subsequences from the alternate sequence for both training and prediction purposes.
The second segment introduces \textbf{Additional Information} into the sequential embeddings, capturing inter-sequence dependencies through an autoregressive transformer. This component predicts responses to each question in the sequence, employing Binary Cross Entropy (BCE) loss calculated based on the prediction outcomes and response label sequences.
The third and final segment conducts \textbf{Auxiliary Task} and focuses on compelling the question embeddings to assimilate skill information. This is achieved by predicting the skill distribution for each question and subsequently minimizing the Kullback–Leibler divergence with the actual skill distribution.}

\subsection{Problem Formulation}
The notations needed to formulate the target of knowledge tracing task and the principle of AAKT are listed in Table \ref{table:notations}. 

\begin{table}[!htbp]
	\caption{{Notations Used for Problem Formulation}}
	\label{table:notations}
	\begin{center}
		{  
			\begin{tabular}{|l|l|} 
				\hline
				Notation & Description \\
				\hline
				$\mathcal{Q}=\{q_1,q_2,\dots\}$ & Question set  \\
                $\mathcal{T}=\{t_1,t_2,\dots\}$ & Skill set  \\
				\hline
                $q_i$ & The i-th question of $\mathcal{Q}$\\
                $t_i$ & The i-th skill of $\mathcal{T}$\\
                \hline
                $\mathcal{I}=\{\mathrm{Ex}_1,\mathrm{Ex}_2,\dots\}$ & The exercise sequence of a student\\
                $\tilde{q}_i\in\mathcal{Q}$ & The question in $\mathcal{I}$ at step $i$\\
                $\tilde{t}_i\subset\mathcal{T}$ & The skill \textbf{set} in $\mathcal{I}$ at step $i$\\
                $\tilde{a}_i\in\mathbb{R}$ & The posterior information in $\mathcal{I}$ at step $i$\\
                $\tilde{c_i}\in \{0,1\}$ & The correctness in $\mathcal{I}$ at step $i$\\
                $\mathrm{Ex}_j=(\tilde{q}_j,\tilde{t}_j,\tilde{a}_j,\tilde{c}_j)\in\mathcal{I}$ & The element of $\mathcal{I}$ at step $j$\\
                \hline
                $L_{max}$ & {Max sequence length}\\
                $r_o\in[0,1]$ & {Overlap ratio}\\
                $L_{batch}$ & {Batch size}\\
                \hline
			\end{tabular}
		}
	\end{center}
\end{table}

Given the exercise sequence $\mathcal{I}$ of a certain student, knowledge tracing seeks to monitor the student's evolving knowledge status by predicting the correctness of future questions. Therefore, suppose that the model has already known the data of $\mathcal{I}'=\{(\tilde{q}_1,\tilde{t}_1,\tilde{a}_1,\tilde{c}_1),\dots,(\tilde{q}_{m},\tilde{t}_{m},\tilde{a}_{m},\tilde{c}_{m})\}\in\mathcal{I}$, it should then predict the probability of answering the next question correctly, namely $\mathrm{Pr}\{\tilde{c}_{m+1}=1\ |\ \mathcal{I}',\tilde{q}_{m+1},\tilde{t}_{m+1}\}$ ($\tilde{a}_{m+1}$ is posterior information so we do not know it before the question is answered).

\subsection{Preliminaries}
\rmk{As presented in the first contribution in Section \ref{sec:intro}, the generative perspective of knowledge tracing task aligns with the setting of autoregressive modeling. We initiate the detailed interpretation by introducing the concept of autoregressive models.

Mathematically, the autoregressive model characterizes a system in which its status (dependent variable) linearly depends on its past status. This system can be mathematically expressed through a stochastic difference equation, as illustrated below:

\begin{equation}
\label{eq:ar}
    y_t=\beta_0+\sum_{i=1}^{p}\beta_i y_{t-i}+\epsilon_t.
\end{equation}

In this equation, $\beta_i$ is a constant denoting how the status $i$ steps ago influences current values. Typically, the expectation is for $\beta_i$ to decrease as $i$ increases, signifying that events from further in the past exert less impact on current occurrences. If anything beyond $p$ steps ago holds no influence, the model is denoted as $AR(p)$. Here, $\epsilon_t$ represents a ``noise'' term accounting for random events affecting the system's status. Actually, the principle of BKT \cite{BKT} is consistent with AR(1), where current values are merely influenced by the status one step ago. In BKT, a student's knowledge status related to a specific skill is decided by the latest response and other fixed probabilities, such as the probability of making a mistake when applying a known skill.}

In deep learning models, autoregressive models (AR models) are mainly used in natural language processing, such as GPT series \cite{GPT1}, \cite{GPT2}, \cite{GPT4}. Unlike original AR models, nonlinear transformation is introduced as a special extension (RNN, Transformer, etc.) to enable more flexible contextual interactions. AR models are unidirectional and can only see preceding information when processing specific sequences. Additionally, the coefficients $\beta_i$ in Eq. \ref{eq:ar} are not constant; instead, they dynamically depend on the model's inputs. Through learning from previous steps and incorporating prior outputs as inputs, AR models recurrently predict the subsequent step. This adaptability positions AR models to excel in both prediction and generative tasks.

When processing sequential information, AR models predict the next step by analyzing previous words and estimating the probability distribution of the whole corpus provided. Formally, if the corpus $\mathcal{C}$ and a sequence $\boldsymbol{x}=\{x_1,x_2,\dots,x_n\} (\forall i\in \{1,\dots,n\},\ x_i\in\mathcal{C})$ is given, the target of AR models can be described as:
\begin{equation}
\begin{aligned}
    &\underset{\Theta}{max}\ \mathcal{L}(\boldsymbol{x}) \\
    =& \mathrm{log} ~
    \mathrm{Pr}(x_1|\Theta)
    \mathrm{Pr}(x_2|x_1;\Theta)\dotsm
    \mathrm{Pr}(x_n|\boldsymbol{x}_{<n};\Theta)\\
    =& \sum_{i=1}^n \mathrm{log}\ \mathrm{Pr}(x_i|\boldsymbol{x}_{<i};\Theta)\\
    =& \sum_{i=1}^n \mathrm{log}\ \frac
    {\mathrm{exp}(\mathrm{hidden}_\Theta(\boldsymbol{x}_{<i})^{\intercal}\cdot\mathrm{emb}(x_i))}
    {\sum_{x\in\mathcal{C}}\mathrm{exp}(\mathrm{hidden}_\Theta(\boldsymbol{x}_{<i})^{\intercal}\cdot\mathrm{emb}(x))},
\end{aligned}
\end{equation}
where $\mathrm{hidden}_\Theta(\cdot)$ and $\mathrm{emb}(\cdot)$ represent hidden states of model and embedding vector respectively.

Similarly, exercise sequences within knowledge tracing tasks embody causal relationships, where the response to the current question hinges on preceding questions and responses. Consequently, we posit that autoregressive models can be effectively employed in knowledge tracing tasks.

\subsection{Alternate Autoregressive Modeling}
\label{sub:alternate}
\subsubsection{Alternate Sequence Construction}
{Some current KT models split the raw datasets into two kinds of sequences, namely history sequence and query sequence \cite{SAKT}, \cite{ATKT}, \cite{KQN}. In accordance with the notations outlined in Table \ref{table:notations}, the exercise sequence $\mathcal{I}=\{\mathrm{Ex}_1,\mathrm{Ex}_2,\dots\}$ is divided and transferred into \textbf{History Sequence} and \textbf{Query Sequence}:
\begin{equation}
\label{eq:3}
\begin{aligned}
    \mathcal{H}_i
    &=\{\mathrm{h}_1,\dots,\mathrm{h}_{i}\}\\
    &=\{\boldsymbol{f}_h(\tilde{q}_1,\tilde{t}_1,\tilde{c}_1),\dots,\boldsymbol{f}_h(\tilde{q}_i,\tilde{t}_i,\tilde{c}_i)\},\\
    \mathcal{Q}_i
    &=\{\mathrm{qry}_1,\dots,\mathrm{qry}_i\}\\
    &=\{\boldsymbol{f}_{qry}(\tilde{q}_1,\tilde{t}_1),\dots,\boldsymbol{f}_{qry}(\tilde{q}_i,\tilde{t}_i)\}.\\
\end{aligned}
\end{equation}}

\begin{figure}[ht]
\centering
\subfigure[{Modeling strategy with knowledge states and question representations.}]{
    \label{fig:autoreg1}
    \includegraphics[width=0.45\textwidth]{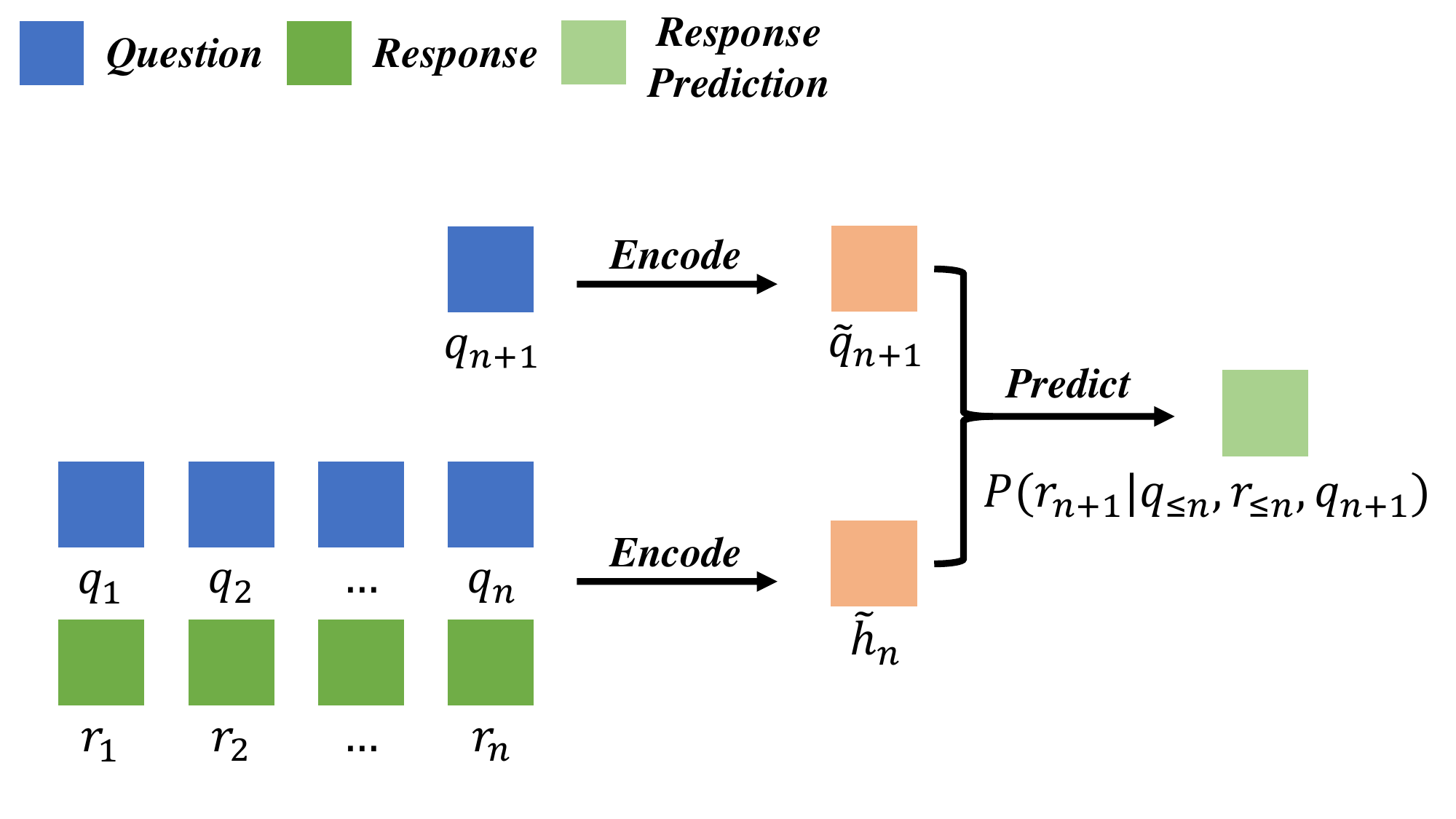}
}
\subfigure[{Modeling strategy with knowledge states and question-integrated knowledge states.}]{
    \label{fig:autoreg2}
    \includegraphics[width=0.45\textwidth]{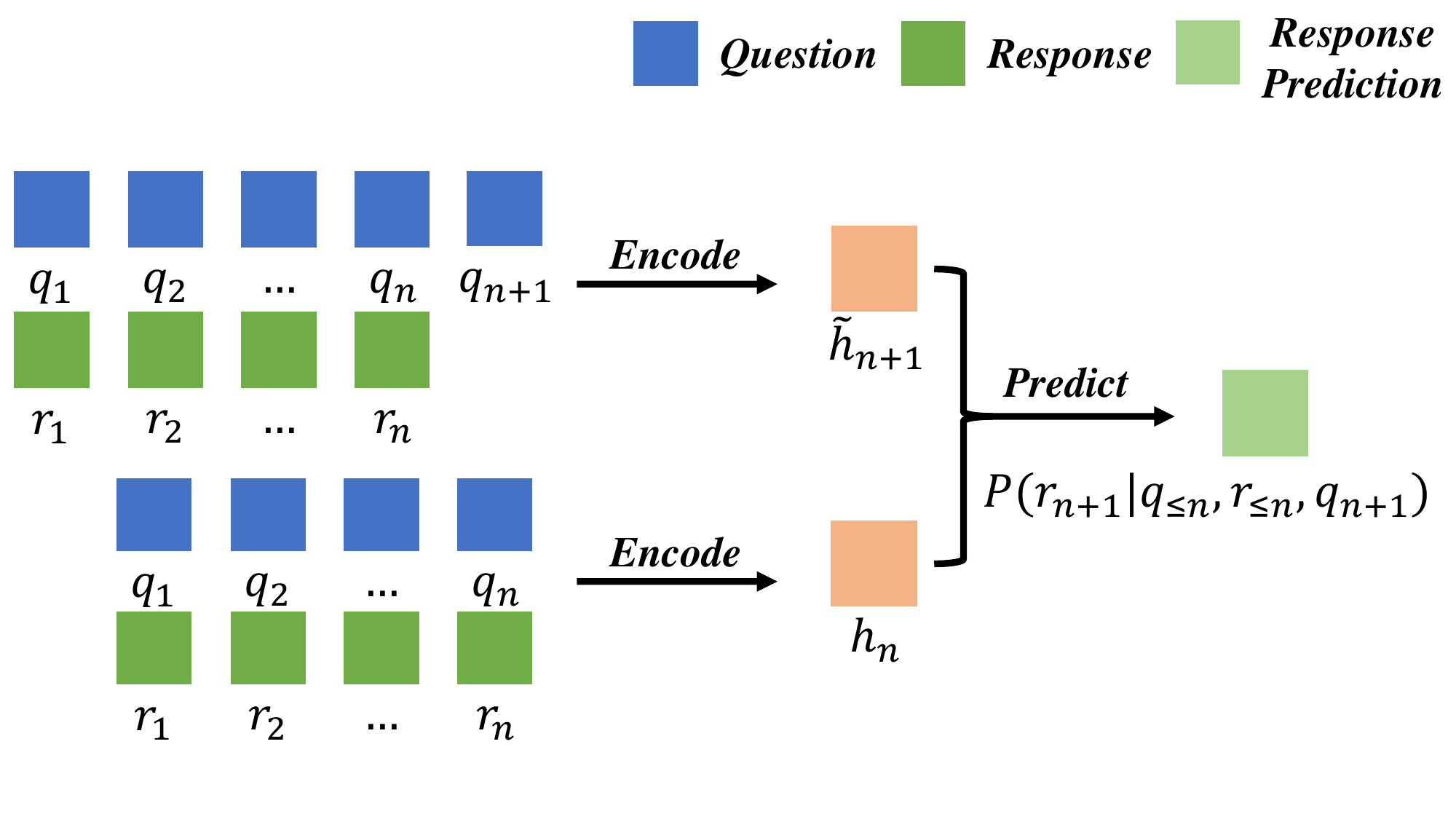}
}
\subfigure[{Modeling strategy with uniform alternating states through alternating sequence.}]{
    \label{fig:autoreg3}
    \includegraphics[width=0.45\textwidth]{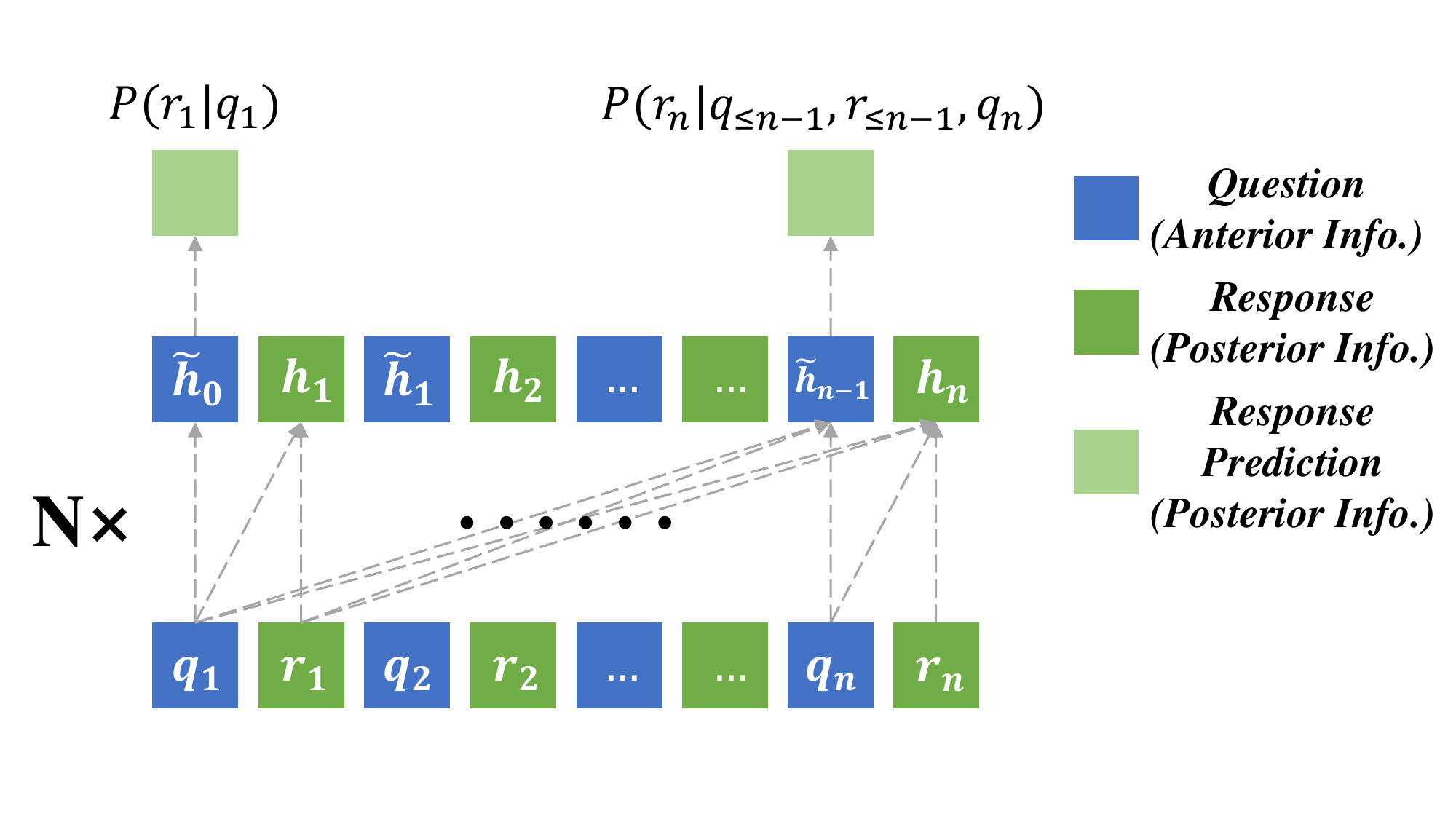}
}
\caption{{Different modeling strategy between previous KT models and AAKT.}}
\label{fig:autoreg}
\end{figure}

\rmk{In Eq. \ref{eq:3}, $\boldsymbol{f}_h(\cdot)$ and $\boldsymbol{f}_{qry}(\cdot)$ denote functions that embed arguments into a vector. history sequence takes every (question, skills, response) tuple in the history interactions, which is the full sequence of student’s interactions in the learning process, and transform every tuple into a vector. Query sequence takes every (question, skills) tuple and does similar transformation. The differences between them lie in the choice of tuple and the transformations. These models learn from students' history exercise sequences and transfer them into hidden knowledge status of every step first. Subsequently, the next element in query sequence is combined with history sequence, generating a predictive result for the correctness of answering the next exercise:}
\begin{equation}
\begin{aligned}
    \hat{{c}}_{i+1}=\mathrm{Predict}(\mathcal{H}_i\,,\,\mathrm{qry}_{i+1})
\end{aligned}
\end{equation}
where $\mathrm{Predict}(\cdot)$ denotes prediction layers of models. {Following the guideline of Eq. 3 and Eq. 4, certain previous KT models developed two main kinds of modeling strategy (refer to Fig. \ref{fig:autoreg1} and Fig. \ref{fig:autoreg2}).}

{Knowledge tracing models like DKT \cite{DKT}, DKT+ \cite{DKT+}, and DTransformer \cite{DTrKT} adopt the approach illustrated in Fig. \ref{fig:autoreg1}, explicitly representing students' status as $\tilde{h}_n$. However, this method requires two separate encoders and introduces a formal disparity between $q_{n+1}$ and $(q_{\leq n},r_{\leq n})$, which can detrimentally impact learning efficiency and generalization.}

{To address the above-mentioned issues, CL4KT \cite{CL4KT} and LBKT \cite{LBKT} utilize the method demonstrated in Fig. \ref{fig:autoreg2}, simultaneously considering knowledge states and question-integrated knowledge states. Here, the models capture students' knowledge states after completing $n$ exercises and when encountering the $(n+1)$th question. While this method mitigates generalization issues, it imposes computational overhead and introduces inconsistency by encoding the same information (i.e., $q_{\leq n}$ and $r_{\leq n}$) twice.}

\begin{figure*}
\subfigure[Sliding Window of Training Dataset]{
    \label{fig:sw1}
    \includegraphics[width=0.45\textwidth]{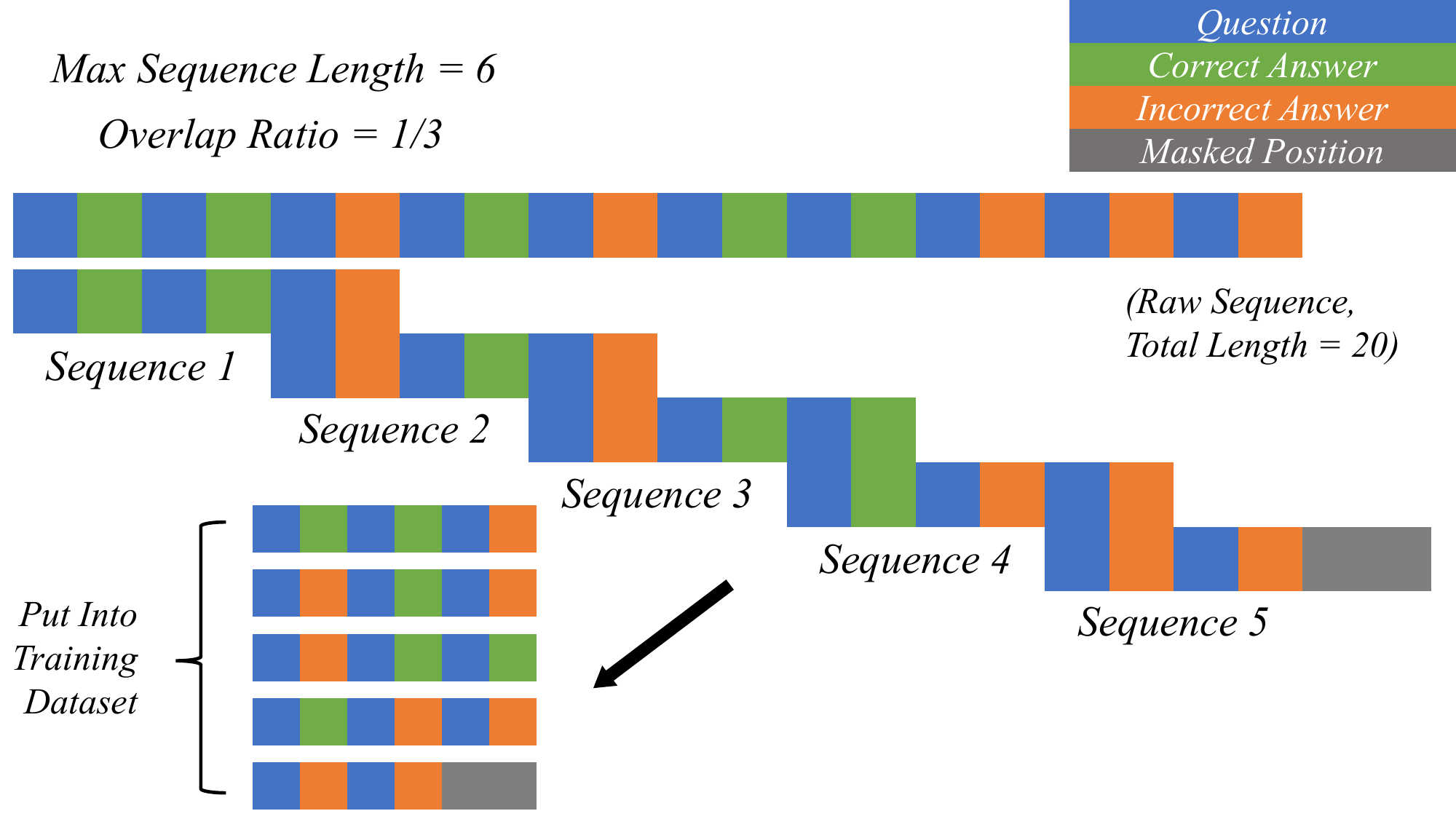}
}
\subfigure[Sliding Window of Testing Dataset (Calculating Metrics)]{
    \label{fig:sw2}
    \includegraphics[width=0.45\textwidth]{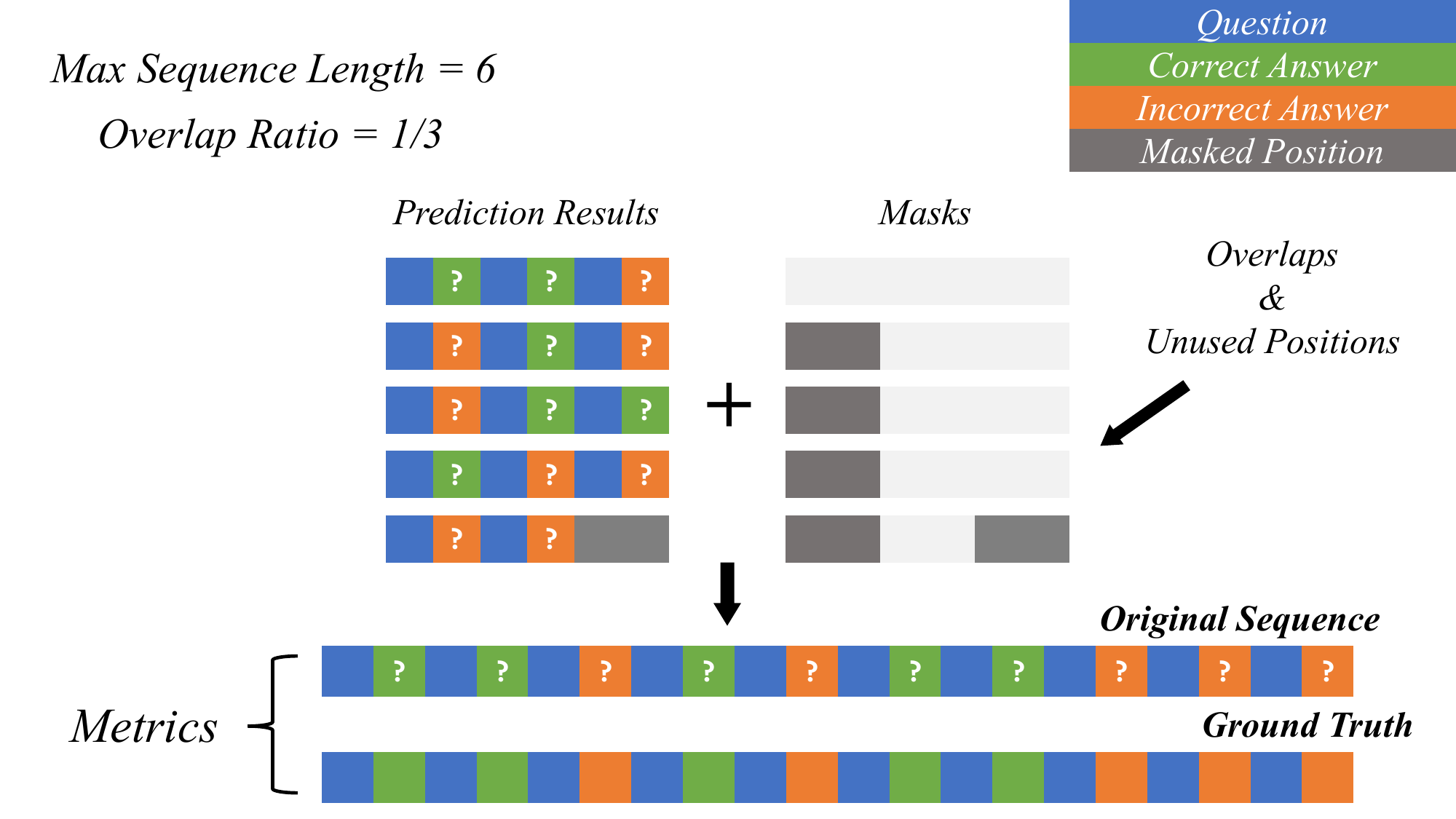}
}
\caption{Overall illustration of sliding window technique in our model. There are slight variations in the approach for the training dataset compared to the testing dataset.}
\label{fig: sw}
\end{figure*}

{Considering information interaction and computational efficiency, we draw inspiration from autoregressive models and introduce a novel modification to the original exercise sequences (refer to Fig. \ref{fig:autoreg3}). The expected result of the previous two strategies is achieved by merely changing the representation of students' exercise records.} The input sequence is formulated as follows, with notations borrowed from Table \ref{table:notations}:
\begin{equation}
\label{eq:our embedding}
\begin{aligned}
    \mathcal{I}^{in}_{i}=
    \begin{cases}
        \boldsymbol{f}_r(\tilde{q}_k,\tilde{t}_k) & ,i=2k-1\\
        \boldsymbol{f}_q(\tilde{c}_k,\tilde{a}_k) & ,i=2k
    \end{cases}(1\leq k\leq \frac{|\mathcal{I}|}{2}).
\end{aligned}
\end{equation}

The embedding functions for questions ($\boldsymbol{f}_q(\cdot)$) and responses ($\boldsymbol{f}_r(\cdot)$) represent the transformation of information related to exercises and skills, and correctness and posterior information, respectively. As evident from Eq. \ref{eq:our embedding}, each element in the original sequences undergoes embedding only once. Consequently, the input sequence utilizes a unified embedding space, thereby enhancing information consistency. \textbf{A comprehensive elucidation of $\boldsymbol{f}_q(\cdot)$ and $\boldsymbol{f}_r(\cdot)$ is provided in Subsection \ref{subsection:auxiliary} and Subsection \ref{subsection:additional}.}

% Moreover, this kind of representation guarantees the full interaction between questions and responses in the processing of autoregressive models (refer to Fig. \ref{fig:autoreg}). At step $i=2k-1$ (where $\mathcal{I}^{in}_i=\boldsymbol{f}_r(\tilde{q}_k,\tilde{t}_k)$), the autoregressive transformer will see every previous steps (namely $\mathcal{I}^{in}_{\leq 2k-1}$ including the current step), therefore generating the prediction of hidden knowledge status at step $i+1=2k$ (where $\mathcal{I}^{in}_{i+1}=\boldsymbol{f}_r(\tilde{c}_k,\tilde{a}_k)$) which is transformed into prediction of correctness afterward. However, the prediction of step $2k+1$ (where $\mathcal{I}^{in}_{2k+1}=\boldsymbol{f}_r(\tilde{q}_{k+1},\tilde{t}_{k+1})$) is useless because it is meaningless for the KT model to predict the next question provided by online exercise platforms or other unrelated sources. Therefore, these useless prediction results are excluded from the final output:
According to the description of Fig. \ref{fig:autoreg3} and Eq. \ref{eq:our embedding}, our model only outputs the response prediction at step $2k-1$. Consequently, the output sequence should be formulated as follows:
\begin{equation}
\begin{aligned}
    \mathcal{I}^{out}_i
    &=\hat{{c}}_i\\
    &=\mathrm{Pr}(\tilde{c}_{i}=1\,|\,\mathcal{I}^{in}_{\leq 2i-1}).
\end{aligned}
\end{equation}

To elaborate further, a single step of hidden knowledge status (represented by a vector with multiple dimensions) is transformed into a 2D vector $\mathbf{v}=(v_{correct},v_{incorrect})$ through a feed-forward network. Here, we interpret the first and second dimensions of $\mathbf{v}$ as the 'raw probability' of predicting correctly and incorrectly, respectively. Then the probability is calculated as follows:
\begin{equation}
\begin{aligned}
    \mathrm{Pr}(\tilde{c}_k=1\,|\,\mathcal{I}^{in}_{\leq 2k-1})
    &=\mathrm{Sigmoid}(v_{correct}-v_{incorrect})\\
    &=\frac{1}{1+\mathrm{exp}(v_{incorrect}-v_{correct})}.
\end{aligned}
\end{equation}

Finally, we use binary cross-entropy loss as the prediction loss:
\begin{equation}
\begin{aligned}
    \mathcal{L}_{pred}=
    -\frac{1}{|\mathcal{I}|}
    \sum_{i=1}^{|\mathcal{I}|}
    \hat{c}_i\cdot \mathrm{log}(\hat{c}_i)+
    (1-\hat{c}_i)\cdot \mathrm{log}(1-\hat{c}_i).
\end{aligned}
\end{equation}

\subsubsection{Sliding Window Technique}
\label{subsection:slidingwindow}
{To optimize information utilization within the datasets, we employ the sliding window technique during preprocessing, permitting overlaps in both the training and testing datasets. However, owing to considerations of validity in the testing process and metric calculations, there are slight variations in the specific steps for training and testing. The sliding window technique is illustrated in Fig. \ref{fig: sw}. In this depiction, two key hyperparameters are set: \textbf{max sequence length} $L_{max}$, representing the maximum length in a data batch, and \textbf{overlap ratio} $r_o$, denoting the extent to which the next sliding window covers the current one (refer to Table \ref{table:notations}).}

For the training dataset, an individual student's alternate exercise sequence (refer to Eq. \ref{eq:our embedding}) is partitioned into subsequences. The window initiates its movement from the commencement of the raw sequence and progresses towards the end. At each step, it moves forward by $L_{max}\cdot (1-r_{o})$ elements, where $L_{max}$ and $r_{o}$ represent the maximum sequence length and overlap ratio, respectively. Additionally, the value of $L_{max}\cdot (1-r_{o})$ must be even, as a complete interaction involves two elements in the sequence (question and answer). If the range of the last window extends beyond the raw sequence, the remainder is masked, ensuring it is not considered when computing the loss. This technique effectively augments the dataset volume by introducing overlaps:
\begin{equation}
    |\mathrm{Dataset}'|\approx\frac{1}{1-r_{o}}|\mathrm{Dataset}|.
\end{equation}

Handling the testing dataset and computing metrics involves some additional processes. In testing, all overlaps in the alternate sequence must be masked at corresponding positions (refer to Fig. \ref{fig:sw2}). This precaution is taken to prevent the same interaction from being counted multiple times during metric calculations, thereby avoiding inconsistency and inaccuracies. The efficacy of the sliding window technique in testing has been validated and shown to enhance prediction performance (refer to Section \ref{section:experiments}).

\subsection{Auxiliary Task in Embedding}
\label{subsection:auxiliary}
To effectively integrate information from both questions and related skills, we introduce an auxiliary task in our model. We operate under the assumption that questions and skills exhibit hierarchical differences and should be treated distinctively. Furthermore, we acknowledge that the information related to questions is more concrete than that of skills and should hold a predominant position in the model.

Specifically, given a question \(q_i \in \mathcal{Q}\), we first transform it into a vector through an embedding matrix \(\mathrm{W}^{q} \in \mathbb{R}^{|\mathcal{Q}| \times dim}\):
\begin{equation}
    \mathrm{emb}(q_i) = \mathrm{W}^{q}_{i} \in \mathbb{R}^{dim}
\end{equation}
where $dim$ denotes the dimension of embedding vectors, and \(\mathrm{W}^{q}_{i}\) denotes the \(i^\text{th}\) row of \(\mathrm{W}^{q}\).

On that basis, we predict the skill distribution beside this question through a classifier for \(\mathcal{T}-\)class classification, which is composed of FFN followed by a softmax function:
\begin{equation}
\tilde{p}(t|q_i)=\mathrm{Softmax}(\mathrm{FNN}(\mathrm{emb}(q_i))), t \in \mathcal{T}
\end{equation}

Simultaneously, we transform \(\tilde{t}_i\), the true skill set of \(q_i\), into the skill distribution for supervision:
\begin{equation}
    p(t|q_i)=
    \begin{cases}
        \frac{1}{|\tilde{t}_i|} & ,t\in \tilde{t}_i \\
        0 & ,t\notin \tilde{t}_i
    \end{cases}
\end{equation}

Finally, we force the question representation $\mathrm{emb}(q_i)$ to learn the information of corresponding skill set through minimizing the KL divergence between the predicted skill distribution \(\tilde{p}(t|q_i)\) and the true skill distribution \(p(t|q_i)\), which forms the auxiliary loss.
\begin{equation}
    \mathcal{L}_{aux}=KL(p(t|q_i)\|\tilde{p}(t|q_i))
\end{equation}

{The efficacy of the auxiliary task in facilitating the integration of skill-related information into the embedding vectors of questions lies in the backpropagation process, where the auxiliary loss is propagated back to the embedding weights of questions. To uphold the hierarchical nature of information, we establish a distinct pathway for skill-related information rather than adding it directly to the embedding vectors of questions. This approach represents both a conceptual and empirical improvement \textbf{(refer to Section \ref{section:experiments})}.}

The auxiliary loss is regarded as one part of the total loss of the model. Therefore, to jointly learn all parameters in our model, the following loss is optimized:
\begin{equation}
\begin{aligned}
    \mathcal{L}=\mathcal{L}_{pred}+\mathcal{L}_{aux}.
\end{aligned}
\end{equation}

\subsection{Processing of Additional Information}
\label{subsection:additional}
\begin{figure}[!htbp]
\centering
\includegraphics[width=\columnwidth]{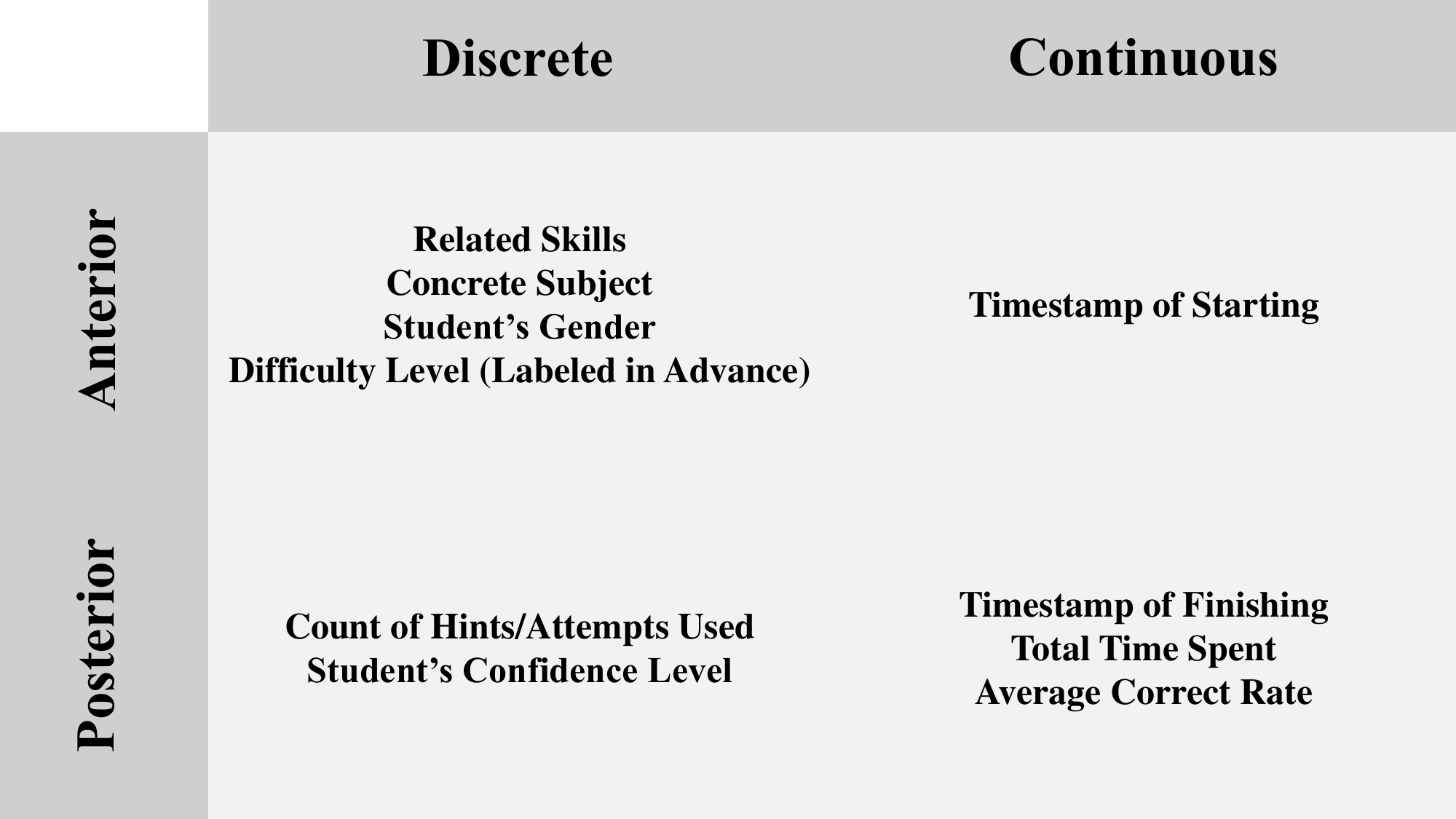}
\caption{Four categories of interaction-related information and examples.}
\label{fig:categories}
\end{figure}

In the context of online education platforms, interaction records encompass diverse attributes, including timestamps and student gender. Considering these attributes in knowledge tracing tasks can be advantageous, as relying solely on question indices and skill sets may inadequately capture the nuances of provided exercises. In addition to semantic information, where some knowledge tracing models leverage textual question descriptions \cite{qDKT}, \cite{SemanticKT}, interaction-related information can be categorized into four distinct groups, as illustrated in Fig. \ref{fig:categories}. 

{In the AAKT framework, we introduce a viable approach to incorporate additional information, both anterior and posterior, into alternate sequences. As illustrated in Fig. \ref{fig:autoreg}, a comprehensive representation of a specific interaction involves two elements in the input sequence. Essentially, anterior information should be combined with $q_i$, while posterior information should be combined with $r_i$ at step $i$.}

In our model, we have focused on utilizing one category: posterior continuous information, specifically the 'total time spent'. This choice is informed by the significance of time spent on exercises in the context of online education and the nature of the datasets. The total time spent values range from 0 ms to approximately $4 \times 10^5$ ms. To bring them into a standardized range (0 ms to $2 \times 10^5$ ms), we applied clipping. Additionally, a constant called \textbf{Time Factor} ($\tau=60000$ ms) was introduced to partially normalize these values:
\begin{equation}
    \mathrm{time}_{i}^{norm}=\frac{\mathrm{time}_i}{\tau}.
\end{equation}

After normalization, these values become dimensionless and can be integrated into the model. More precisely, the normalized time values undergo multiplication by a trainable vector, denoted as $\mathbf{v}_{time}$, followed by addition to the embedding vectors of $r_i$ (refer to Eq. \ref{eq:our embedding}):
\begin{equation}
\begin{aligned}
    \tilde{a}_i&=\mathrm{time}_i^{norm},\\
    \boldsymbol{f}_q(\tilde{a}_i,\tilde{c}_i)&=
    \begin{cases}
        \tilde{a}_i \cdot \mathbf{v}_{time}+\mathrm{emb}_{right} & ,\tilde{c}_i=1\\
        \tilde{a}_i \cdot \mathbf{v}_{time}+\mathrm{emb}_{wrong} & ,\tilde{c}_i=0
    \end{cases}.
\end{aligned}
\end{equation}

Here, $emb_{right}$ and $emb_{wrong}$ are two trainable vectors.

\begin{table*}[htbp]
	\caption{{Datasets Statistics}}
	\label{table:datasets}
	\begin{center}
		{  
			\begin{tabular}{lcccc}
				\hline
				Features/Datasets & EdNet-KT1 & {ASSISTments2009} & ASSISTments2017 & Junyi \\
				\hline
                \#Students & 5,000 & {4,151} & 1,709 & 247,605\\
                \#Questions & 12,276 & {16,891} & 3,162 & 721\\
                \#Skills & 188 & {110} & 99 & 40\\
                \#Records & 2,208,254 & {274,042} & 942,540 & 25,925,983\\
                \#Avg. Questions per Skill & 65.30 & {153.55} & 31.94 & 18.02\\
                \#Avg. Skills per Question & 2.14 & {1.19} & 1.00 & 1.00\\
                \hline
                \%Correct Answering Records & 69.88\% & {66.15\%} & 37.27\% & 82.79\%\\
                \%Incorrect Answering Records & 30.12\% & {33.85\%} & 62.73\% & 17.21\%\\
                \hline
                \%Sequences within Length (0,10) & 0.00\% & {55.36\%} & 0.00\% & 49.56\%\\
                \%Sequences within Length [10,100) & 23.75\% & {30.76\%} & 3.63\% & 33.50\%\\
                \%Sequences within Length [100,1000) & 53.03\% & {13.13\%} & 82.67\% & 14.80\%\\
                \%Sequences within Length [1000,10000) & 22.06\% & {0.75\%} & 13.70\% & 2.13\%\\
                \%Sequences within Length [10000,+$\infty$) & 1.16\% & {0.00\%} & 0.00\% & 0.01\%\\
                \hline
			\end{tabular}
		}
	\end{center}
\end{table*}

\begin{figure*}
    \centering
    \subfigure{\includegraphics[width=0.24\textwidth]{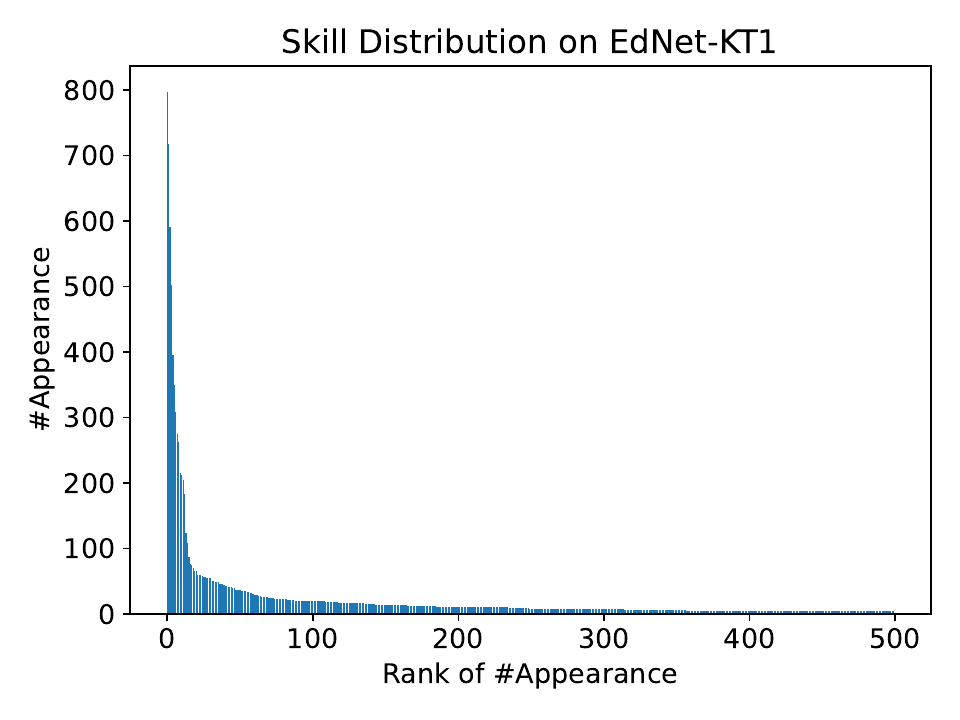}}
    \subfigure{\includegraphics[width=0.24\textwidth]{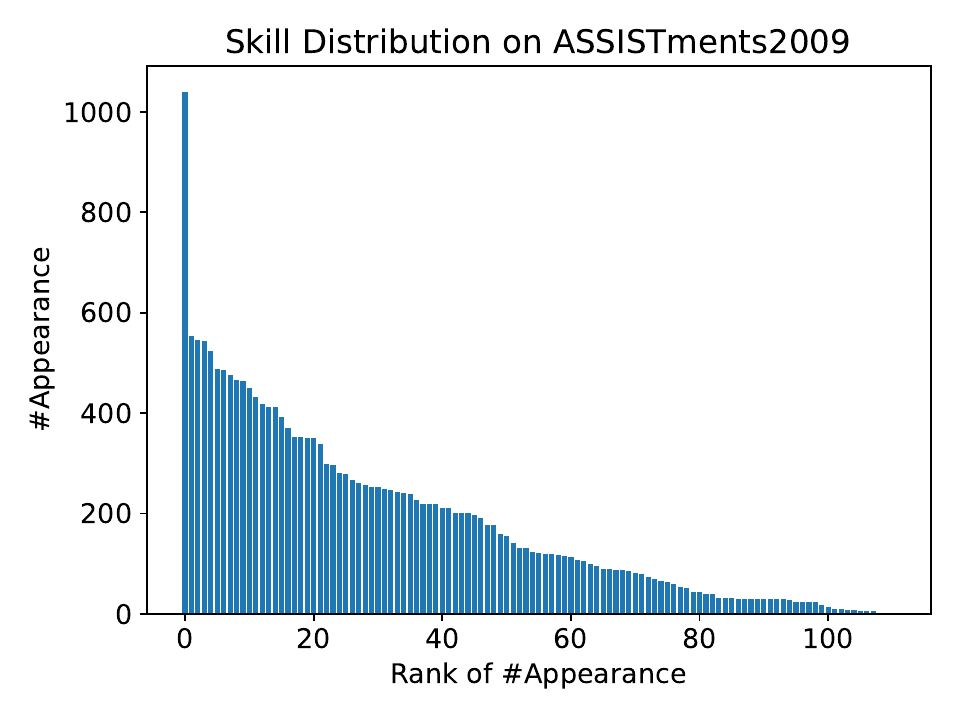}}
    \subfigure{\includegraphics[width=0.24\textwidth]{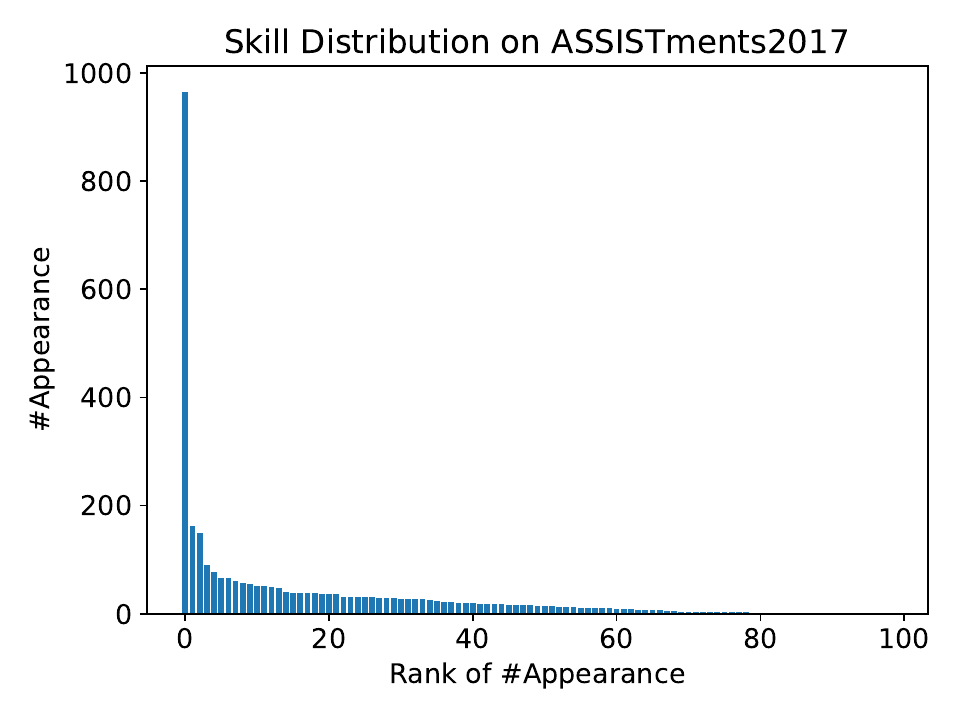}}
    \subfigure{\includegraphics[width=0.24\textwidth]{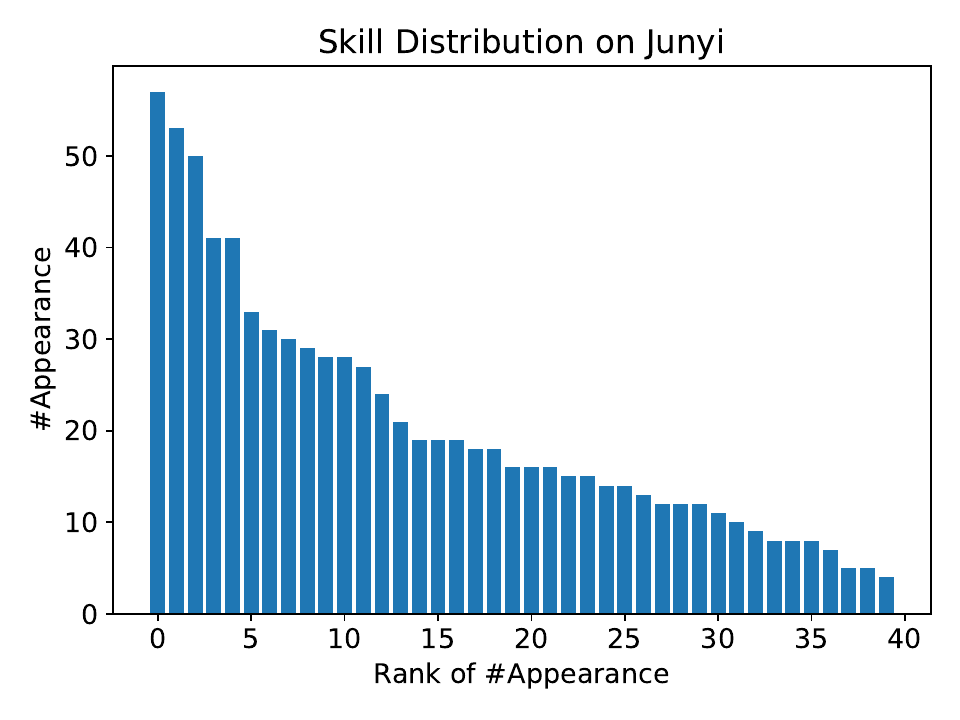}}
    \caption{{Skill distribution on every dataset. For EdNet-KT1, we define that two questions are the same in terms of skill if and only if their skill sets are identical. Moreover, considering the extremely long and thin tail of skill distribution on EdNet-KT1, we truncate the list to 500 records.}}
    \label{fig:dist}
\end{figure*}

\section{Experiments}
\label{section:experiments}
In this section we evaluate the proposed model's performance through a series of experiments on real-world datasets\footnote{The code is available at https://github.com/alxzzhou/AAKT}. The ablation studies and visualizations are also conducted to further validate the effectiveness of specific components of AAKT.

\subsection{Datasets}
To evaluate the performance of AAKT, we utilize four real-world datasets, each containing information on the total time spent on every question. The statistics of four datasets are shown in Table \ref{table:datasets} and Fig. \ref{fig:dist}. 

\subsubsection{EdNet-KT1}
{This dataset\footnote{This dataset is available at: https://github.com/riiid/ednet} was collected and introduced by \cite{ednet}. Due to the excessively large number of records, we selected the EdNet-KT1 dataset, which consists of students' practice history logs, and randomly extracted the records of 5,000 students. The dataset includes a total of 188 skills and 12,276 questions, with the maximum number of concepts underlying a single question being 6.}

\subsubsection{ASSISTments2009}
{This dataset, provided by the ASSISTments platform \cite{assistment2009}, is one of the most popular benchmark datasets in knowledge tracing (KT). In our study, we use the latest updated version, "Skill-builder". This version includes data from 4,151 students, 16,891 questions, and 110 concepts, resulting in a total of 274,042 observed responses. \textbf{Notably, questions with more than one skill are split into multiple questions in the original dataset, which is not proper. To address this issue, we combine them as one question in the experiments.}}

\subsubsection{ASSISTments2017}
{The dataset\footnote{This dataset is available at: https://sites.google.com/view/\\assistmentsdatamining/data-mining-competition-2017} is sourced from the ASSISTments Data Mining Competition 2017. We applied the same preprocessing methods as those used for ASSISTments2009.}

\subsubsection{Junyi}
{\textbf{JunyiAcademy}\footnote{https://www.junyiacademy.org/} \cite{junyi} collects data on JunyiAcademy platform from 2018/08 to 2019/09. After preprocessing, the dataset has a total of 721 exercises and 25,925,983 interactions, with 247,605 users.}

\subsection{Baselines and Evaluation Metrics}
We compare AAKT with the following advanced KT models. {Specifically, we classify the baseline into two categories according to the modeling strategy illustrated in Fig. \ref{fig:autoreg}.}

{Baselines using knowledge states and question representations:}
\begin{itemize}
    \item \textbf{DKT} \cite{DKT}, the first knowledge tracing model to leverage deep neural networks, employs a long short-term memory recurrent neural network (LSTM) for sequential prediction of student performance. By extracting the hidden state of the LSTM, the model can deduce the student's mastery level across various skills.
    \item \textbf{DKT+} \cite{DKT+} is an improvement of DKT. It adds a regularization term on the hidden states of students, smoothing the evolving of knowledge status.  
    \item \textbf{DKVMN} \cite{DKVMN} extends DKT by introducing extra memory-augmented neural networks. Two matrices (namely key matrix and value matrix) are used to better trace students' hidden knowledge status. Moreover, DKVMN utilizes the 'read' and 'write' processes to emulate the process of learning.
    \item \textbf{SAKT} \cite{SAKT} is the initial knowledge tracing model to incorporate a self-attention mechanism. Its overall architecture closely mirrors that of the transformer.
    \item \textbf{AKT} \cite{AKT} models the influence of history interactions on students' current learning status by adopting an innovative monotonic attention. It also introduces the difficulty coefficient when encoding skills.
    \item \textbf{KQN} \cite{KQN} introduces the concept of probabilistic skill similarity, which links specific metrics of distance between skill vectors to the odds ratios of skills.
    \item \textbf{ATKT} \cite{ATKT} utilizes adversarial training combined with an attentive LSTM to model students' knowledge status. The training process of the model contains perturbance to improve robustness. 
    \item \textbf{CL4KT} \cite{CL4KT} is an innovative KT model for its contrastive learning framework. It is designed to uncover students' learning histories with similar/dissimilar semantic information, thereby facilitating model's understanding of specific relationships.
    \item \textbf{DTransformer} \cite{DTrKT} addresses the challenge of knowledge state tracing through the Diagnostic Transformer architecture. Additionally, it proposes a novel training paradigm which aligns with contrastive learning and emphasizes the stability of knowledge diagnosis.
\end{itemize}

{Baselines using knowledge states and question-integrated knowledge states:}
\begin{itemize}
    \item \textbf{CoKT} \cite{CoKT} utilizes inter-student information by identifying peer students with similar question-answering experiences. It integrates the inter-student information with intra-student data to effectively trace knowledge states and predict students' correctness in answering questions.
    \item \textbf{LBKT} \cite{LBKT} enhances prediction results by combining the forgetting factor with learners' knowledge acquisition, allowing for a comprehensive update of their dynamic knowledge states. This method achieves superior results compared to existing approaches.
\end{itemize}

This paper employs the \textbf{Area Under the ROC Curve} (AUC) as one of the metrics for assessing model performance, a standard measure in knowledge tracing tasks \cite{DKT, DKVMN, AKT}. An AUC value closer to 1 indicates superior predictive performance for students' responses, while a value of 0.5 reflects the effectiveness of random guessing. Moreover, \textbf{Accuracy} (ACC) and \textbf{Root Mean Square Error} (RMSE) are included where higher ACC and lower RMSE indicate better performance.

\subsection{Implementation}
\subsubsection{Details of Autoregressive Transformer}
In our model, we utilize GPT-J \cite{GPT-J} instance (without pre-trained weights) as the autoregressive transformer. The design of GPT-J closely follows GPT-3 Curie \cite{GPT3}. However, two minor architectural improvements have been made:
\begin{itemize}
    \item Rotary Positional Embedding (RoPE) \cite{RoPE} is adopted for better performance.
    \item The attention layer and the feedforward layer are placed in parallel for decreased communication.
\end{itemize}

\subsubsection{Other Details}
We conduct 5-fold cross-validation for each combination of models and datasets. For training, we use 80\% of the student sequences, while the remaining 20\% are utilized for model evaluation.

Our AAKT model is implemented with PyTorch and we use Adam optimizer to train our model, setting the learning rate to 0.001 for all datasets. All experiments are conducted on a server with one NVIDIA A5000 (24GB) graphic card. Balancing time cost and performance, we set the overlap ratio $r_{o}$ to 0.5 both in training and testing. Moreover, the implementations of baselines are based on open-source code provided by corresponding authors and pyKT \cite{pykt}, a comprehensive python-based benchmark platform for knowledge tracing tasks.

\begin{table*}[ht] 
	\caption{{The Performance of AAKT and All Baselines on 4 Datasets}}
	\label{table:result}
	\begin{center} 
		{   
			\begin{tabular}{l|ccc|ccc|ccc|ccc}
				\hline
				    & \multicolumn{3}{c|}{EdNet-KT1} & \multicolumn{3}{c|}{ASSISTments2009} & \multicolumn{3}{c|}{ASSISTments2017} & \multicolumn{3}{c}{Junyi} \\
                
                        & AUC $\uparrow$ & ACC $\uparrow$ & RMSE $\downarrow$ & AUC $\uparrow$ & ACC $\uparrow$ & RMSE $\downarrow$ & AUC $\uparrow$ & ACC $\uparrow$ & RMSE $\downarrow$ & AUC $\uparrow$ & ACC $\uparrow$ & RMSE $\downarrow$\\
                \hline
                DKT     & 0.7609 & 0.7410 & 0.4161 & 0.6785 & 0.6839 & 0.4541 & 0.7615 & 0.7058 & 0.4349 & 0.8029 & 0.8543 & 0.3380 \\
                DKT+    & 0.7640 & 0.7435 & 0.4140 & 0.6829 & 0.6823 & 0.4548 & 0.7758 & 0.7092 & 0.4331 & 0.8025 & \underline{0.8590} & 0.3305 \\
                DKVMN   & 0.7589 & 0.7380 & 0.4212 & 0.6471 & 0.6670 & 0.4722 & 0.7317 & 0.6895 & 0.4441 & 0.8003 & 0.8522 & 0.3338 \\
                SAKT    & 0.7640 & 0.7426 & 0.4141 & 0.7103 & 0.6633 & 0.4746 & 0.7325 & 0.6935 & 0.4468 & 0.7996 & 0.8529 & 0.3327 \\
                AKT     & 0.7733 & 0.7498 & 0.4124 & 0.7233 & 0.7170 & 0.4369 & 0.7711 & 0.7181 & 0.4282 & 0.8057 & 0.8578 & 0.3313 \\
                KQN     & -       & -       & -       & -       & -       & -       & 0.7067 & 0.6801 & 0.4516 & 0.7560 & 0.8465 & 0.3446 \\
                ATKT    & -       & -       & -       & -       & -       & -       & 0.7214 & 0.6932 & 0.4431 & 0.7597 & 0.8499 & 0.3418 \\
                CL4KT   & 0.7741 & 0.7489 & 0.4103 & 0.7152 & 0.7142 & 0.4371 & 0.7786 & 0.7239 & 0.4247 & 0.8070 & 0.8561 & 0.3308 \\
                DTransformer  
                        & \underline{0.7797} & \underline{0.7539} & \underline{0.4076} & \underline{0.7303} & \underline{0.7193} & \underline{0.4345} & \underline{0.7922} & \underline{0.7320} & \underline{0.4214} & 0.8104 & {0.8582} & \underline{0.3285} \\
                \hline
                CoKT    & 0.7771 & 0.7507 & 0.4091 & 0.7246 & 0.7161 & 0.4365 & 0.7840 & 0.7271 & 0.4241 & 0.8098 & 0.8570 & 0.3298 \\
                LBKT    & -       & -       & -       & 0.7285 & 0.7191 & 0.4348 & 0.7883 & 0.7302 & 0.4239 & \underline{0.8126} & 0.8583 & 0.3311 \\
                \hline
                AAKT (Ours) 
                        & \textbf{0.7827} & \textbf{0.7554} & \textbf{0.4064} & \textbf{0.7357} & \textbf{0.7223} & \textbf{0.4340} & \textbf{0.8018} & \textbf{0.7361} & \textbf{0.4198} & \textbf{0.8146} & \textbf{0.8603} & \textbf{0.3281} \\
                \hline
			\end{tabular}
		}
	\end{center} 
\end{table*}

\begin{table*}[ht] 
	\caption{The Performance of AAKT and All Baselines on Filtered ASSISTments2009 and Junyi}
	\label{table:remove10}
	\begin{center}
		{  
			\begin{tabular}{l|ccc|ccc}
				\hline
				        & \multicolumn{3}{c|}{ASSISTments2009 (Filtered)} & \multicolumn{3}{c}{Junyi (Filtered)} \\
                            & AUC $\uparrow$ & ACC $\uparrow$ & RMSE $\downarrow$ & AUC $\uparrow$ & ACC $\uparrow$ & RMSE $\downarrow$\\
				\hline
                DKT         & 0.6830 & 0.6755 & 0.4579  & 0.8030 & 0.8476 & 0.3311 \\
                DKT+        & 0.6861 & 0.6787 & 0.4570  & 0.8021 & 0.8524 & 0.3309 \\
                DKVMN       & 0.6499 & 0.6578 & 0.4788  & 0.7937 & 0.8499 & 0.3340 \\
                SAKT        & 0.6627 & 0.6601 & 0.4750  & 0.8020 & 0.8531 & 0.3326 \\
                AKT         & 0.7189 & 0.7087 & 0.4391  & 0.8087 & 0.8548 & 0.3347 \\
                KQN         & -      & -      & -       & 0.7606 & 0.8422 & 0.3421 \\
                ATKT        & -      & -      & -       & 0.7650 & 0.8423 & 0.3438 \\
                CL4KT       & 0.7181 & 0.7164 & 0.4380  & 0.8024 & 0.8532 & 0.3319 \\
                DTransformer& \underline{0.7338} & 0.7185 & \underline{0.4340} & 0.8138 & 0.8617 & \underline{0.3258} \\
                \hline
                CoKT        & 0.7269 & 0.7190 & 0.4345  & 0.8104 & 0.8590 & 0.3291 \\
                LBKT        & 0.7322 & \underline{0.7210} & 0.4342 & \underline{0.8144} & \underline{0.8620} & 0.3270 \\
                \hline
                AAKT (Ours) & \textbf{0.7370} & \textbf{0.7266} & \textbf{0.4338} & \textbf{0.8179} & \textbf{0.8644} & \textbf{0.3237} \\
                \hline
			\end{tabular}
		}
	\end{center}
\end{table*}

\begin{table*}[ht]
	\caption{The Correlation Coefficients between Different Metrics}
	\label{table:corrcoef}
	\begin{center}
		{  
			\begin{tabular}{l|cccc}
				\hline
			                     & EdNet-KT1 & ASSISTments2009 & ASSISTments2017 & Junyi \\
				\hline
                ACC and AUC       & 0.98965 & 0.80560& 0.97162&0.88353\\
                AUC and RMSE      & -0.95895& -0.98810 & -0.99085&-0.91221\\
                ACC and RMSE      & -0.93551& -0.76554&  -0.97392&-0.92088\\
                \hline
			\end{tabular}
		}
	\end{center}
\end{table*}

\rmk{
Details of other hyperparameters are listed below. For hyperparameters that are not fixed, we utilized grid searching to find the optimal choice.
\begin{itemize}
    \item {Batch size is chosen from \{32, 64, 128, 256\}.}
    \item Dimension of embedding is chosen from $\{64,96,128,256\}$.
    \item Max sequence length is chosen from $\{20,50,100,300,500,1000,2000\}$ according to the distribution of sequences in each dataset.
    \item The number of heads in multi-head attention is set to 8. This configuration is borrowed from vanilla Transformer \cite{Attention}.
    \item {The number of GPT-J blocks is chosen from \{2,3,4\}.}
    \item The dimension of rotary position embedding is set to $\frac{d_{emb}}{2n_{head}}$ where $d_{emb}$ and $n_{head}$ denotes embedding dimension and number of heads in multi-head attention respectively. This configuration is borrowed from the GitHub repository \textbf{ChatGLM-Finetuning} \footnote{https://github.com/liucongg/ChatGLM-Finetuning/blob/\\a137ad63ee5c7d4e2c020e6f7f12d1bf71320bd1/glm3/modeling\_chatglm.py}.
\end{itemize}
}

\begin{table*}[htbp]
	\caption{The Performance of Original AAKT Model and Its Variants}
	\label{table:ablation}
	\begin{center}
		{  
                \scriptsize
			\begin{tabular}{l|ccc|ccc|ccc|ccc}
				\hline
				    & \multicolumn{3}{c|}{EdNet-KT1} & \multicolumn{3}{c|}{ASSISTments2009} & \multicolumn{3}{c|}{ASSISTments2017} & \multicolumn{3}{c}{Junyi} \\
                
                        & AUC $\uparrow$ & ACC $\uparrow$ & RMSE $\downarrow$ & AUC $\uparrow$ & ACC $\uparrow$ & RMSE $\downarrow$ & AUC $\uparrow$ & ACC $\uparrow$ & RMSE $\downarrow$ & AUC $\uparrow$ & ACC $\uparrow$ & RMSE $\downarrow$\\
                \hline
                AAKT                        & \textbf{0.7827} & \textbf{0.7554} & \textbf{0.4064} & \textbf{0.7357} & \textbf{0.7223} & \textbf{0.4340} & \textbf{0.8018} & \textbf{0.7361} & \textbf{0.4198} & \textbf{0.8146} & \textbf{0.8603} & \textbf{0.3281} \\
				\hline
                w/o Sliding            & 0.7693 & 0.7500 & 0.4081 & 0.7302 & 0.7141 & 0.4385 & 0.7827 & 0.7241 & 0.4288 & 0.7998 & 0.8531 & 0.3305 \\
                w/o Skill              & 0.7713 & 0.7515 & 0.4076 & 0.7309 & 0.7160 & 0.4378 & 0.7974 & 0.7310 & 0.4260 & 0.8086 & 0.8563 & 0.3293 \\
                w/o Time               & 0.7726 & 0.7531 & 0.4072 & 0.7315 & 0.7181 & 0.4372 & 0.7846 & 0.7287 & 0.4248 & 0.8094 & 0.8568 & 0.3291 \\
                w/o Time\&Skill        & 0.7690 & 0.7485 & 0.4083 & 0.7299 & 0.7139 & 0.4381 & 0.7819 & 0.7254 & 0.4281 & 0.8046 & 0.8550 & 0.3301 \\
                w/o Auxiliary Task     & 0.7679 & 0.7477 & 0.4090 & 0.7282 & 0.7134 & 0.4410 & 0.7831 & 0.7267 & 0.4305 & 0.8029 & 0.8502 & 0.3315 \\
                \hline
			\end{tabular}
		}
	\end{center}
\end{table*}

\subsection{Results and Discussions}
In this section, we list the average results of cross-validation in Table \ref{table:result} \textbf{($-$ means that the model does not fit the dataset)}, where the highest results in every column are displayed in bold and the second highest results are underlined.  From the data shown in Table \ref{table:result}, the following observations regarding baselines and AAKT can be made: 
\begin{itemize}
    \item AAKT outperforms the state-of-the-art models on four datasets (i.e., EdNet-KT1, ASSISTments2009, ASSISTments2017, and Junyi) and three metrics, validating its effectiveness in knowledge tracing tasks.
    \item DKT and DKT+ have relatively high performances on EdNet-KT1 and ASSISTments2017, outperforming models with more complicated structures like DKVMN and SAKT. DKT and DKT+ are simple models that merely utilize the LSTM structure. This implies that more skillful techniques should be explored in modeling the learning sequences of students for knowledge tracing and complexity does not necessarily lead to accuracy.
    {\item Transformer-based models in baselines (i.e., AKT and DTransformer) perform well on four datasets. Notably, DTransformer achieves the second highest metric on EdNet-KT1, ASSISTments2009, and ASSISTments2017. The utilization of the diagnostic transformer enables DTransformer to outperform SAKT and AKT which use simpler attention techniques and training paradigms.
    \item Contrastive learning is applicable in knowledge tracing and demonstrates great potential. Models with contrastive learning (i.e., CL4KT and DTransformer) have relatively high performance compared to all chosen baselines.}
\end{itemize}

{Moreover, to further investigate the inter-metric relationship in Table \ref{table:result}, we calculate the correlation coefficient of every metric pair on every dataset (refer to Table \ref{table:corrcoef}). These relationships can guide researchers and practitioners in selecting the most appropriate metrics for assessing model performance, considering the trade-offs between ACC, AUC, and RMSE based on the specific goals of educational knowledge tracing tasks. The following observations regarding different metrics can be made:}

{\begin{itemize}
    \item The correlation coefficients between Accuracy (ACC) and Area Under the Curve (AUC) consistently demonstrate a strong positive linear relationship across all datasets, ranging from 0.80560 to 0.98965. This suggests that models achieving higher accuracy also tend to exhibit higher AUC values. Additionally, the other two pairs of metrics show a negative correlation.
    \item The correlation coefficients demonstrate variations across diverse datasets. Specifically, the correlation coefficient between ACC and AUC on ASSISTments2009 is relatively lower compared to other datasets. This characteristic is also observed in the relationship between ACC and RMSE.
\end{itemize}}

As depicted in Table \ref{table:datasets}, approximately half of the sequences in ASSISTments2009 and Junyi are shorter than 10 steps. Additionally, around 20\% of students in ASSISTments2009 responded to only one question. The prevalence of extremely short sequences in these datasets introduces a potential bias in model predictions, particularly for models sensitive to shorter sequences. To mitigate this influence, sequences shorter than 10 steps in ASSISTments2009 and Junyi are excluded, and the performances of models are reassessed (refer to Table \ref{table:remove10}). This operation is also seen in EKT \cite{EKT}. These processed datasets are marked as \textbf{``Filtered''}.

Generally, the performances on filtered ASSISTments2009 and Junyi show some fluctuation. Our model ranks first on these two datasets.

\begin{figure*}
\centering
    \subfigure[AUC Distribution of Random Batches on ASSISTment2009]{
        \label{fig:a09}
        \includegraphics[width=0.22\textwidth]{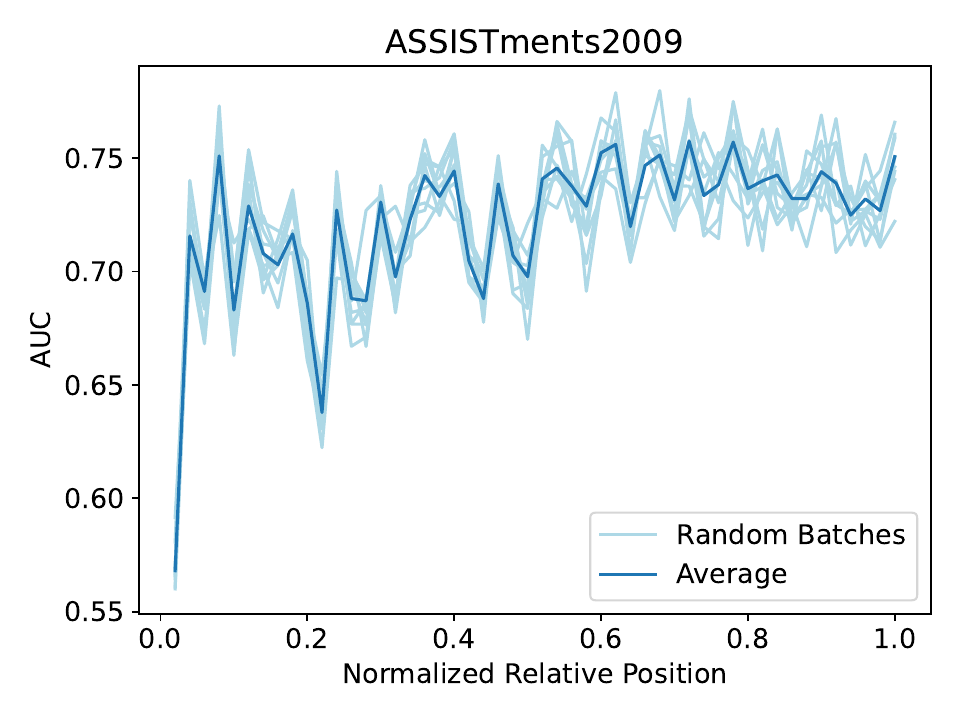}
    }
    \subfigure[AUC Distribution of Random Batches on ASSISTment2017]{
        \label{fig:a17}
        \includegraphics[width=0.22\textwidth]{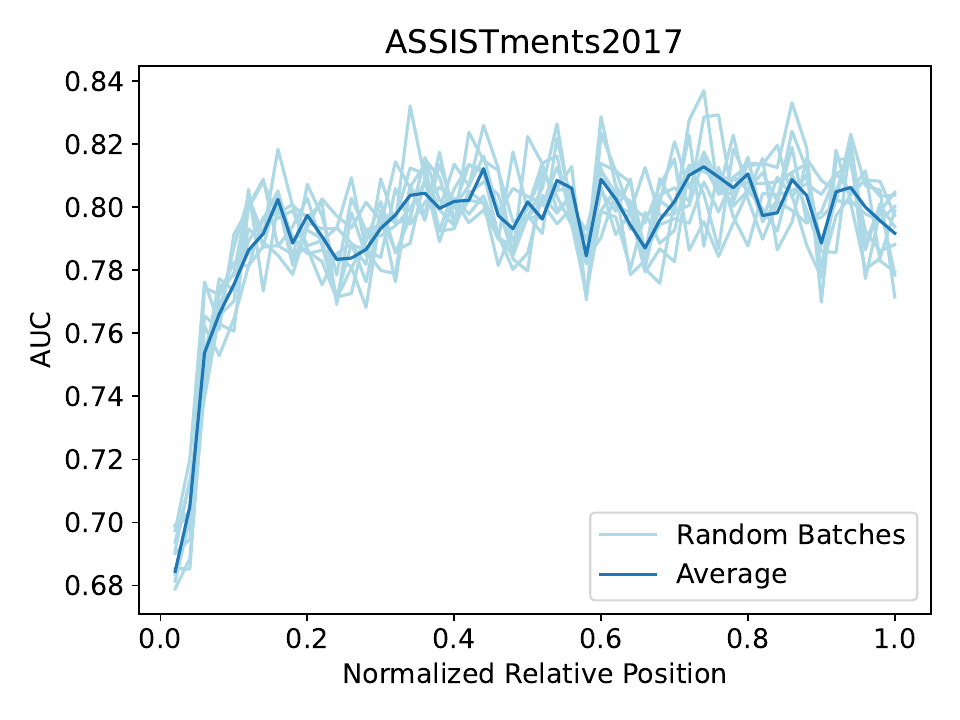}
    }
    \subfigure[AUC Distribution of Random Batches on EdNet-KT1]{
        \label{fig:ednet}
        \includegraphics[width=0.22\textwidth]{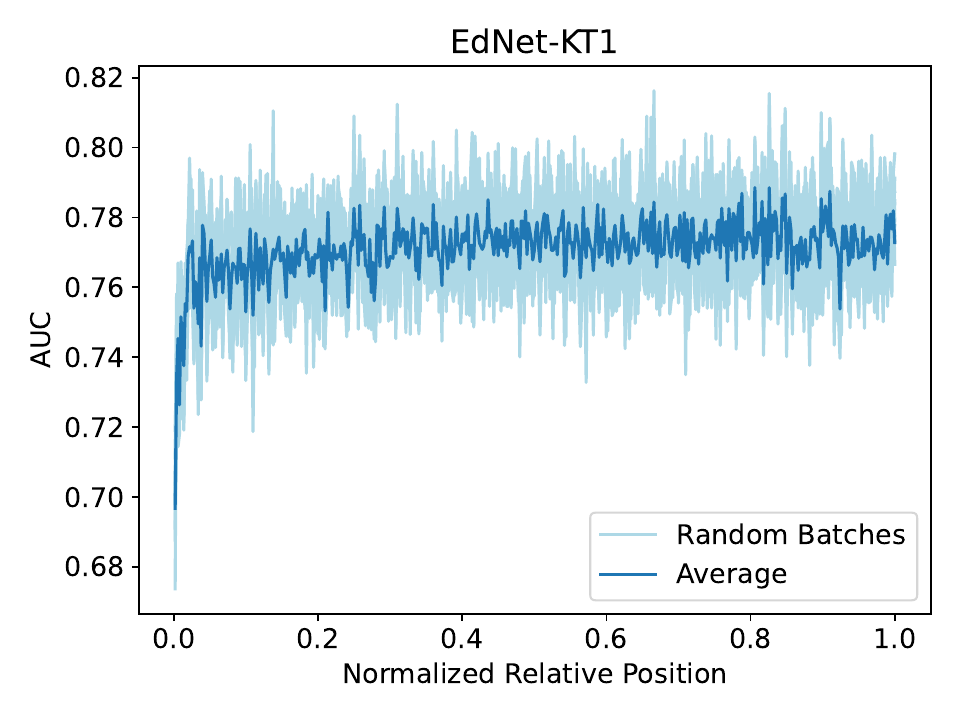}
    }
    \subfigure[AUC Distribution of Random Batches on Junyi]{
        \label{fig:junyi}
        \includegraphics[width=0.22\textwidth]{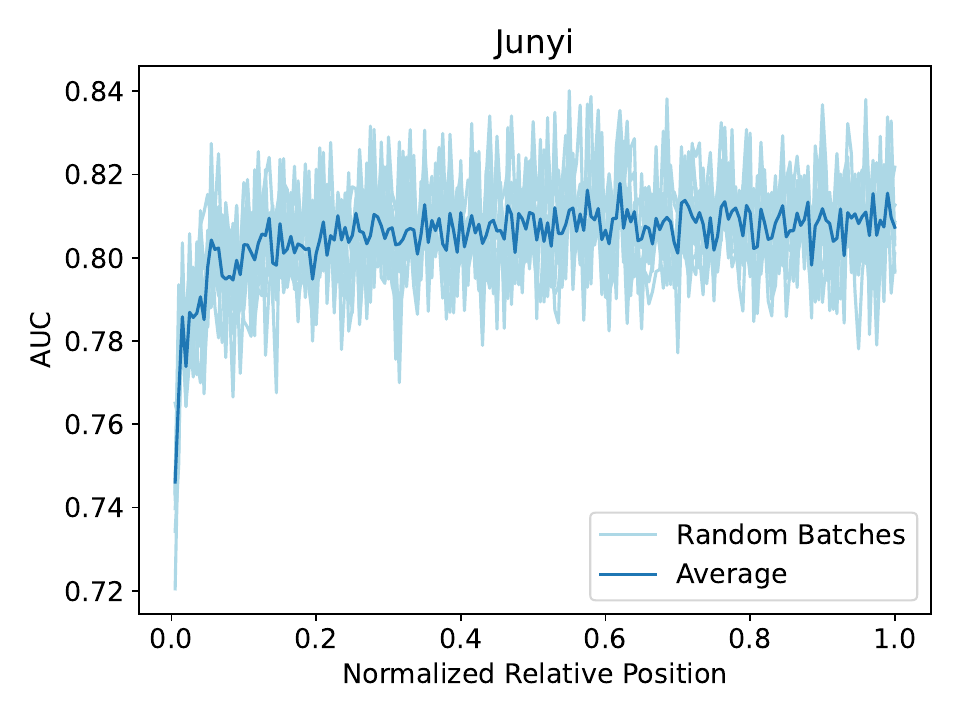}
    } 
    \caption{{AUC distribution on random batches in different datasets. The x-axis denotes the relative position of each question in the sequence. Light blue lines are results of 10 randomly chosen batches in each figure and the dark blue line is their average.}}
    \label{fig:sliding}
\end{figure*}

\subsection{Ablation Study}
To further investigate the contribution of key components in AAKT, we conducted a series of ablation studies. Variants of AAKT with specific components removed are defined. Results are listed in Table \ref{table:ablation}.
\begin{itemize}
    \item \textbf{AAKT w/o Sliding} removes sliding window technique. That is to say, the $r_{o}$ is set to zero both in training and testing.
    \item \textbf{AAKT w/o Skill} removes the information related to skills.
    \item \textbf{AAKT w/o Time} removes the attribute `total time spent' for every question in the dataset.
    \item \textbf{AAKT w/o Time\&Skill} removes both `total time spent' attributes and the information related to skills.
    \item \textbf{AAKT w/o Auxiliary Task} removes the auxiliary task. Specifically, it directly adds the embedding vectors of questions and skills together. For questions with more than one skill, we take the average of skill embedding vectors.
\end{itemize}

From Table \ref{table:ablation}, the following observations can be made:
\begin{itemize}
    \item Removal of any key component (i.e., Sliding Window Technique, Additional Information, and Auxiliary Task) results in a decline in metrics across all four datasets. This underscores the validated effectiveness of these key components in our model.
    \item AAKT without the Sliding Window exhibits inferior performance across all datasets, affirming the positive impact of the Sliding Window technique.
    \item AAKT without Time \& Skill performs less optimally than both AAKT without Time and AAKT without Skill. This observation suggests that incorporating more additional information contributes to improved model performance.
    \item Across all datasets, AAKT without Skill outperforms AAKT without the Auxiliary Task, highlighting the negative influence of an inadequate embedding strategy. The assumption of equal importance between questions and skills in embedding proves less effective within the AAKT framework.
\end{itemize}

\begin{figure}
\centering
    \includegraphics[width=0.9\columnwidth]{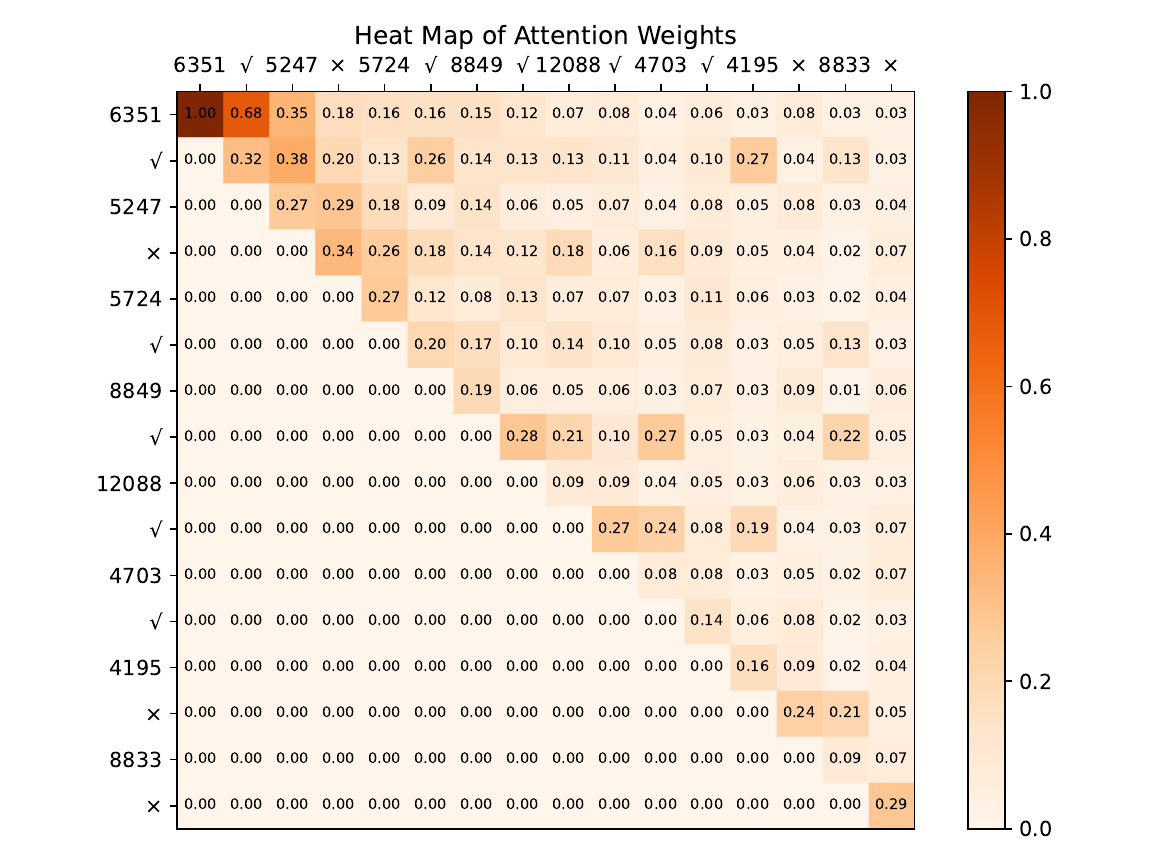}
    \caption{Heat map of attention weights of first causal self-attention layer in our model. The x-axis and y-axis denote the same interaction subsequence from a certain sequence in EdNet-KT1 and the elements denote the specific attention weights (with sotfmax). Our model predicts all the answers to this subsequence correctly. Therefore, attention weights can demonstrate the preference of our model on the history of interactions when predicting.}
    \label{fig:heat}
\end{figure}

\subsection{Visualization}
In this subsection, we conduct some visualizations to get deeper insights into some key components of AAKT.

\subsubsection{Case Study}
In this part, a simple case study is conducted to reinforce the validation of alternate sequence construction and autoregressive modeling. We select a specific alternate sequence from EdNet-KT1 where our model predicted the first 8 answers correctly. Moreover, the attention weights of the first attention layer in our model are extracted when it is calculating the prediction results. A heat map is plotted according to the attention weights (refer to Fig. \ref{fig:heat}).

Our model has different preferences on history interactions when predicting the answer to the next question. Specifically, it emphasizes the result of answering question 5247 when predicting the answer of 5724. Similar emphasis can be observed from the prediction of 12088, 4703, 4195, and 8833. Our model utilizes the information of history responses to predict the answer to the current question, successfully and efficiently combining the information of questions and responses. This result is credited to alternative autoregressive modeling of original sequences.

\subsubsection{The Effectiveness of Sliding Window}
To validate the effectiveness of the sliding window technique, it is imperative to examine alternate sequences horizontally. A meticulous analysis of the model's performance in predicting responses at specific positions is warranted.

Firstly, we demonstrate the overall AUC distribution on different positions and different datasets. We set the value of $r_{o}$ to $0.0$ and choose 10 random batches from the testing process on every dataset and calculate AUC in terms of the position in the sequence. These batches with shape $(L_{batch}, L_{max})$ are transposed to $(L_{max}, L_{batch})$. Then for every position $1\leq i\leq L_{max}$, there is a 1D array recording every prediction in this batch. After removing the masked and useless positions, AUC is calculated on these 1D arrays. Fig. \ref{fig:sliding} shows the visualizations of this experiment (the settings of $L_{max}$ may be different on each dataset). The AUC of starting 10\% prediction results are relatively lower, then increasing quickly. Afterward, the AUC fluctuates randomly and dramatically around a fixed value.

\begin{figure}
    \centering
    \includegraphics[width=0.8\columnwidth]{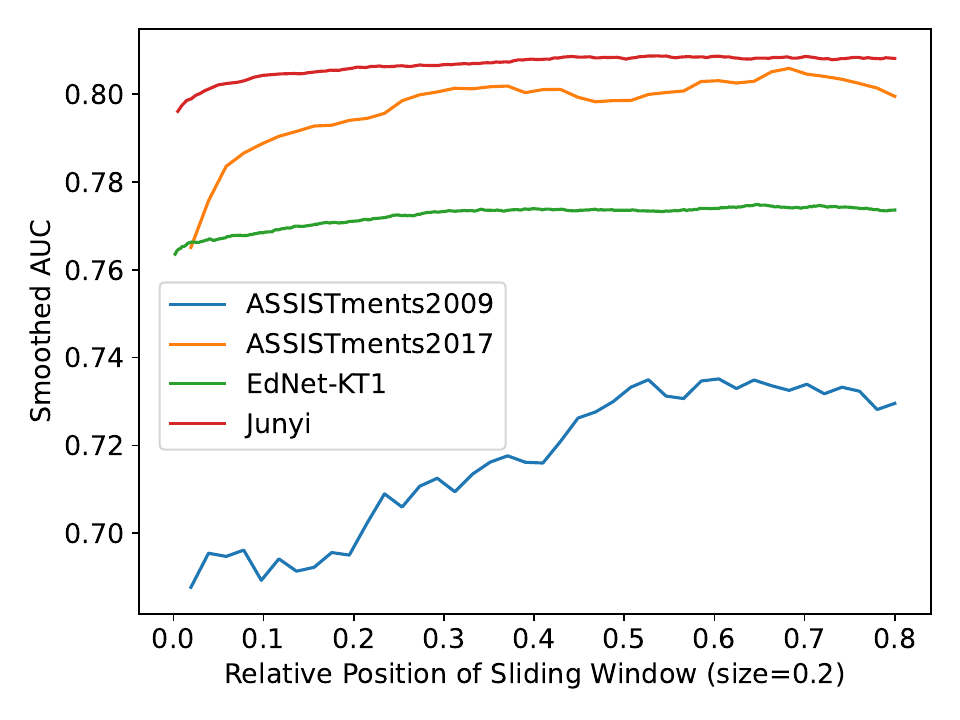}
    \caption{{Smoothed AUC results in Fig. \ref{fig:sliding} using a sliding window with the length of $0.2\times max\_sequence\_length$. The x-axis denotes the normalized relative position of the starting point of the sliding window.}}
    \label{fig:smoothed AUC}
\end{figure}

\begin{figure}
    \centering
    \includegraphics[width=0.8\columnwidth]{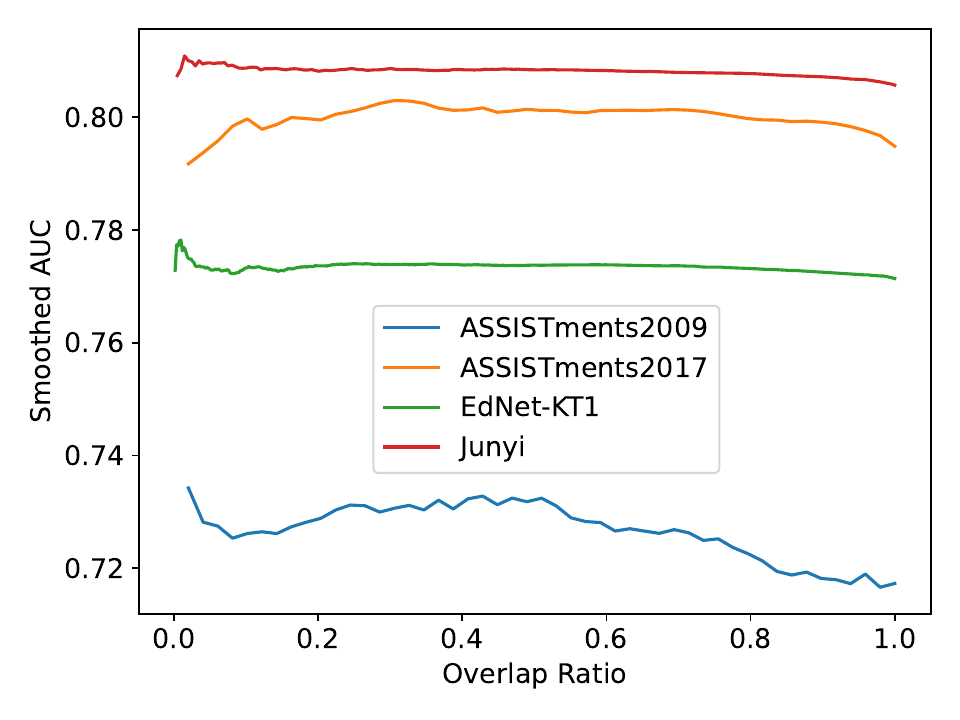}
    \caption{{The relationship between overlap ratio and the mean AUC score in the corresponding sliding window.} }
    \label{fig:length of sliding window}
\end{figure} 

\begin{figure*}
    \centering
    \subfigure{\includegraphics[width=0.4\textwidth]{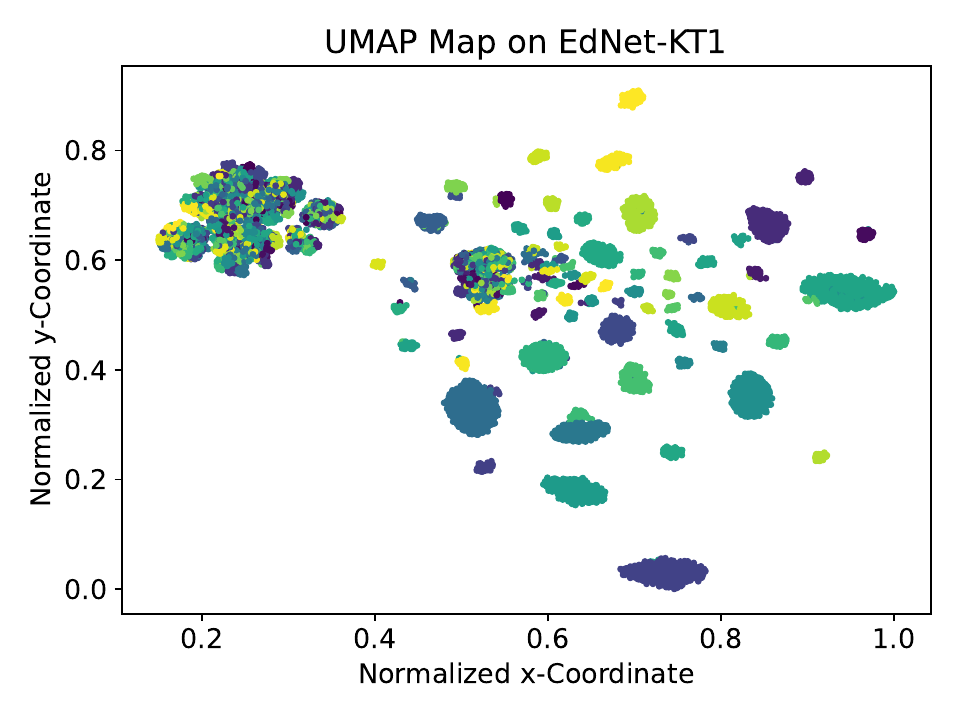}} 
    \subfigure{\includegraphics[width=0.4\textwidth]{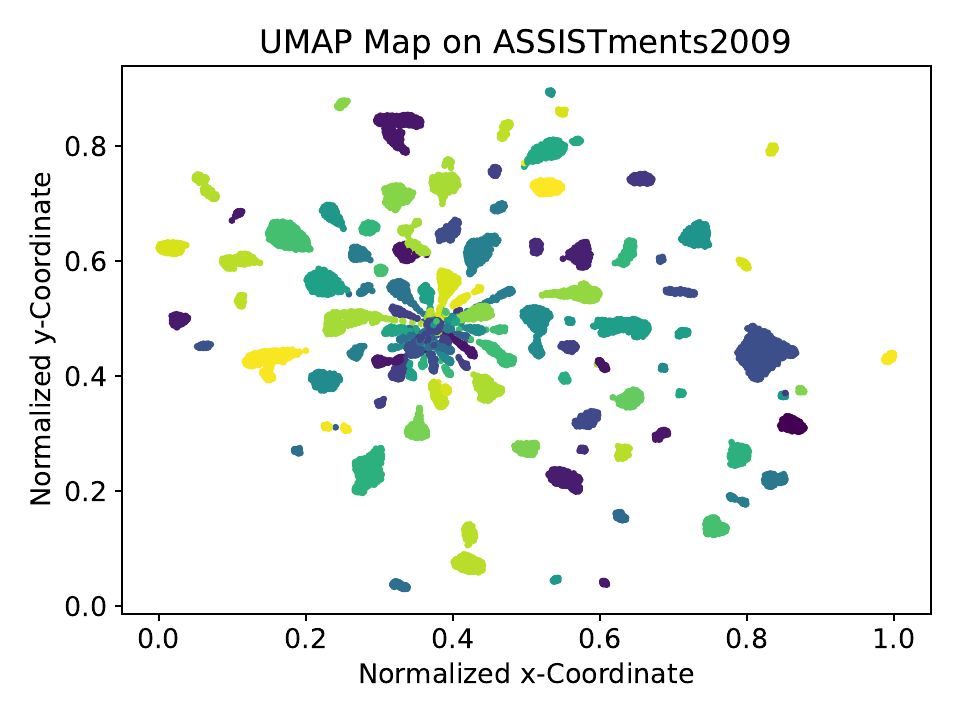}}\vskip 1pt
    \subfigure{\includegraphics[width=0.4\textwidth]{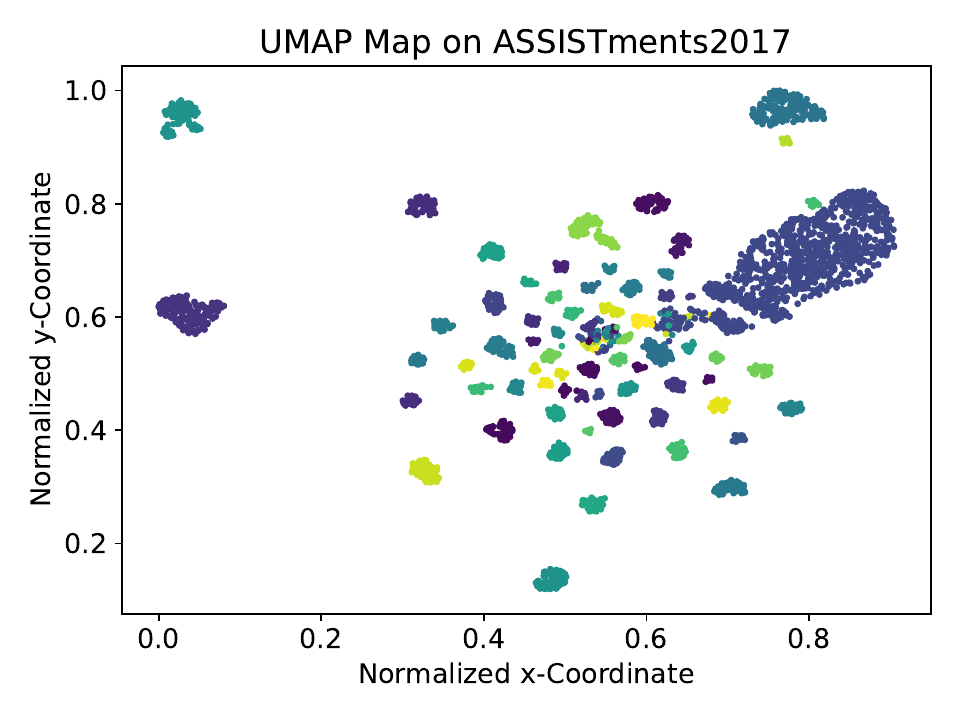}}
    \subfigure{\includegraphics[width=0.4\textwidth]{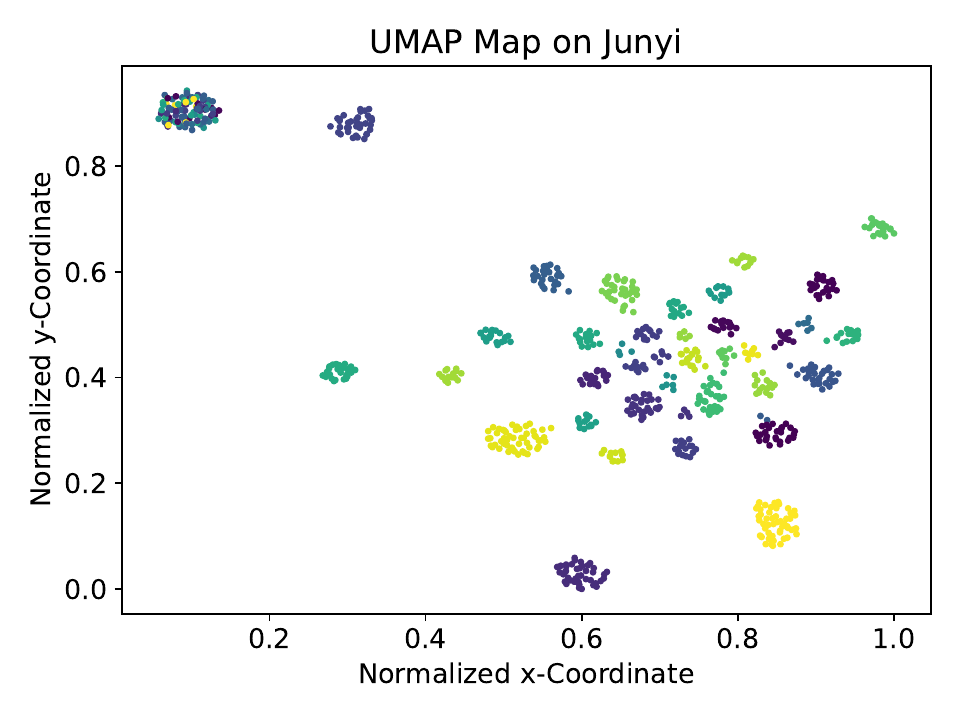}}
    \caption{{UMAP mappings of embedding vectors processed by the auxiliary task. Distinct colors within each subfigure signify different skills or skill sets. It is important to note that distances between clusters hold no explicit meaning.}}
    \label{fig:umap}
\end{figure*}

To make an in-depth analysis of Fig. \ref{fig:sliding}, we smoothed all these results using a sliding window with the length of $0.2\times L_{max}$. In Fig. \ref{fig:smoothed AUC}, the outcomes of the smoothing process are presented, revealing discernible patterns amidst the randomness. Generally, a noticeable upward trend in mean AUC scores is evident as the sliding window progresses through randomly selected batches from EdNet-KT1, Junyi, and ASSISTments2017. Although ASSISTments2009 exhibits more fluctuations, the overall trend persists.

Following the explanation provided in Subsection \ref{subsection:slidingwindow}, as $r_{o}$ decreases from $1.0$, the prediction window expands from the last element of each sequence. Fig. \ref{fig:length of sliding window} illustrates the relationship between $r_{o}$ and the mean AUC in the corresponding prediction window, using the same randomly selected batches of data described above. The line representing ASSISTments2009 exhibits a small peak around 0.5, while others remain relatively stable. Given the information presented in Fig. \ref{fig:length of sliding window}, the fixed value of 0.5 for $r_{o}$ in our model is demonstrated to be a reasonable and close-to-optimal choice.

\subsubsection{The Effectiveness of Auxiliary Task}
To validate whether the auxiliary task effectively incorporates skill-related information into the embedding vectors of questions, we perform an experiment to visualize the overall distribution on a 2D plane using UMAP \cite{UMAP}, a proficient dimension reduction algorithm. The outcomes, depicted in Fig. \ref{fig:umap}, clearly indicate the successful accomplishment of the auxiliary task, as it distinctly segregates embedding vectors of questions based on their respective skill(s) in UMAP maps.

Detailed analysis of UMAP maps on every dataset are listed as follows: 
\begin{itemize}
    \item \textbf{EdNet-KT1}: The majority of embedding vectors are correctly categorized. However, there is a cluster that contains various types of points. This phenomenon can be ascribed to the notably skewed distribution of skills across the question set, as illustrated in Fig. \ref{fig:dist}. The distribution in EdNet-KT1 presents an elongated and slender tail, with certain skill sets appearing only once in the comprehensive question set.
    \item \textbf{ASSISTments2009}: The majority of embedding vectors are correctly categorized. 
    \item \textbf{ASSISTments2017}: The majority of embedding vectors are correctly categorized. 
    \item \textbf{Junyi}: The majority of embedding vectors are correctly categorized. 
\end{itemize}

\section{Conclusion}
In this paper, we developed AAKT, an advanced knowledge tracing model based on the autoregressive transformer. Different from previous knowledge tracing models with transformers which separate question and response, we combine question and response sequences, constructing alternate sequences as input. By treating the knowledge tracing task as a generative process, we utilize the GPT-J instance (without pre-trained weights) as an autoregressive transformer backbone to extract the hidden knowledge status of students. We also introduce the sliding window technique to allow valid overlaps between batches both in training and evaluation, maximizing the utility of limited datasets and enhancing overall performance. To emphasize the hierarchical relationship between questions and skills, we introduce an auxiliary task to better combine them. Moreover, we analyze a paradigm of handling additional information in the setting of our model and select 'total time spent' as additional information. Comparative experiments involving 11 previous knowledge tracing models on four real-world datasets affirm the efficacy of AAKT in knowledge tracing tasks. The results from extensive ablation studies and visualized analysis further validate the effectiveness of key components in AAKT, including the alternate sequence construction, sliding window implementation, and auxiliary task integration.

Possible directions of future work include 1) designing more advanced methodologies for combining information from questions and skills, 2) exploring more efficient autoregressive transformer structures tailored for knowledge tracing tasks, and 3) devising more efficient approaches for utilizing limited datasets in training and testing phases.

\ifCLASSOPTIONcaptionsoff
  \newpage
\fi

\bibliographystyle{IEEEtran}
\bibliography{main}

\begin{thebibliography}{10}
\providecommand{\url}[1]{#1}
\csname url@samestyle\endcsname
\providecommand{\newblock}{\relax}
\providecommand{\bibinfo}[2]{#2}
\providecommand{\BIBentrySTDinterwordspacing}{\spaceskip=0pt\relax}
\providecommand{\BIBentryALTinterwordstretchfactor}{4}
\providecommand{\BIBentryALTinterwordspacing}{\spaceskip=\fontdimen2\font plus
\BIBentryALTinterwordstretchfactor\fontdimen3\font minus \fontdimen4\font\relax}
\providecommand{\BIBforeignlanguage}[2]{{%
\expandafter\ifx\csname l@#1\endcsname\relax
\typeout{** WARNING: IEEEtran.bst: No hyphenation pattern has been}%
\typeout{** loaded for the language `#1'. Using the pattern for}%
\typeout{** the default language instead.}%
\else
\language=\csname l@#1\endcsname
\fi
#2}}
\providecommand{\BIBdecl}{\relax}
\BIBdecl

\bibitem{BKT}
A.~T. Corbett and J.~R. Anderson, ``Knowledge tracing: Modeling the acquisition of procedural knowledge,'' \emph{User Modeling and User-Adapted Interaction}, vol.~4, no.~4, pp. 253--278, 1994.

\bibitem{IndividualizedBKT}
M.~Yudelson, K.~R. Koedinger, and G.~J. Gordon, ``Individualized {Bayesian} knowledge tracing models,'' in \emph{Proceedings of the 16th International Conference on Artificial Intelligence in Education}, 2013, pp. 171--180.

\bibitem{BKTOverview}
R.~Pel{\'{a}}nek, ``Bayesian knowledge tracing, logistic models, and beyond: An overview of learner modeling techniques,'' \emph{User Modeling and User-Adapted Interaction}, vol.~27, no. 3-5, pp. 313--350, 2017.

\bibitem{DKT}
C.~Piech, J.~Bassen, J.~Huang, S.~Ganguli, M.~Sahami, L.~J. Guibas, and J.~Sohl{-}Dickstein, ``Deep knowledge tracing,'' in \emph{Proceedings of the 29th Annual Conference on Neural Information Processing Systems}, 2015, pp. 505--513.

\bibitem{Survey1}
G.~Abdelrahman, Q.~Wang, and B.~P. Nunes, ``Knowledge tracing: {A} survey,'' \emph{ACM Computing Surveys}, vol.~55, no.~11, pp. 224:1--224:37, 2023.

\bibitem{Survey2}
X.~Song, J.~Li, T.~Cai, S.~Yang, T.~Yang, and C.~Liu, ``A survey on deep learning based knowledge tracing,'' \emph{Knowledge-Based Systems}, vol. 258, p. 110036, 2022.

\bibitem{DKT+}
C.~Yeung and D.~Yeung, ``Addressing two problems in deep knowledge tracing via prediction-consistent regularization,'' in \emph{Proceedings of the 5th Annual {ACM} Conference on Learning at Scale}, 2018, pp. 5:1--5:10.

\bibitem{DKVMN}
J.~Zhang, X.~Shi, I.~King, and D.~Yeung, ``Dynamic key-value memory networks for knowledge tracing,'' in \emph{Proceedings of the 26th International Conference on World Wide Web}, 2017, pp. 765--774.

\bibitem{SKVMN}
G.~Abdelrahman and Q.~Wang, ``Knowledge tracing with sequential key-value memory networks,'' in \emph{Proceedings of the 42nd International {ACM} {SIGIR} Conference on Research and Development in Information Retrieval}, 2019, pp. 175--184.

\bibitem{Attention}
A.~Vaswani, N.~Shazeer, N.~Parmar, J.~Uszkoreit, L.~Jones, A.~N. Gomez, L.~Kaiser, and I.~Polosukhin, ``Attention is all you need,'' in \emph{Proceedings of the 31st Annual Conference on Neural Information Processing Systems}, 2017, pp. 5998--6008.

\bibitem{SAKT}
S.~Pandey and G.~Karypis, ``A self attentive model for knowledge tracing,'' in \emph{Proceedings of the 12th International Conference on Educational Data Mining}, 2019.

\bibitem{AKT}
A.~Ghosh, N.~T. Heffernan, and A.~S. Lan, ``Context-aware attentive knowledge tracing,'' in \emph{Proceedings of the 26th {ACM} {SIGKDD} Conference on Knowledge Discovery and Data Mining}, 2020, pp. 2330--2339.

\bibitem{DTrKT}
Y.~Yin, L.~Dai, Z.~Huang, S.~Shen, F.~Wang, Q.~Liu, E.~Chen, and X.~Li, ``Tracing knowledge instead of patterns: Stable knowledge tracing with diagnostic transformer,'' in \emph{Proceedings of the ACM Web Conference 2023}, 2023, pp. 855--864.

\bibitem{SAINT+}
D.~Shin, Y.~Shim, H.~Yu, S.~Lee, B.~Kim, and Y.~Choi, ``{SAINT+:} integrating temporal features for ednet correctness prediction,'' in \emph{Proceedings of the 11th International Learning Analytics and Knowledge Conference}, 2021, pp. 490--496.

\bibitem{VanillaTrKT}
S.~Pu, M.~Yudelson, L.~Ou, and Y.~Huang, ``Deep knowledge tracing with transformers,'' in \emph{Proceedings of the 21st International Conference on Artificial Intelligence in Education}, 2020, pp. 252--256.

\bibitem{ATKT}
X.~Guo, Z.~Huang, J.~Gao, M.~Shang, M.~Shu, and J.~Sun, ``Enhancing knowledge tracing via adversarial training,'' in \emph{Proceedings of the 29th {ACM} Multimedia Conference}, 2021, pp. 367--375.

\bibitem{BKTAssumption1}
R.~S.~J. de~Baker, A.~T. Corbett, and V.~Aleven, ``More accurate student modeling through contextual estimation of slip and guess probabilities in bayesian knowledge tracing,'' in \emph{Proceedings of the 9th International Conference on Intelligent Tutoring Systems}, 2008, pp. 406--415.

\bibitem{BKTAssumption2}
Z.~A. Pardos and N.~T. Heffernan, ``{KT-IDEM}: Introducing item difficulty to the knowledge tracing model,'' in \emph{Proceedings of the 19th International Conference on User Modeling, Adaption and Personalization}, 2011, pp. 243--254.

\bibitem{BKTImprovement1}
M.~Khajah, R.~V. Lindsey, and M.~Mozer, ``How deep is knowledge tracing?'' in \emph{Proceedings of the 9th International Conference on Educational Data Mining}, 2016.

\bibitem{BKTImprovement2}
S.~Shen, Z.~Huang, Q.~Liu, Y.~Su, S.~Wang, and E.~Chen, ``Assessing student's dynamic knowledge state by exploring the question difficulty effect,'' in \emph{Proceedings of the 45th International {ACM} {SIGIR} Conference on Research and Development in Information Retrieval}, 2022, pp. 427--437.

\bibitem{KTM}
J.~Vie and H.~Kashima, ``Knowledge tracing machines: Factorization machines for knowledge tracing,'' in \emph{Proceedings of the 33rd {AAAI} Conference on Artificial Intelligence, the 31st Innovative Applications of Artificial Intelligence Conference, the 9th {AAAI} Symposium on Educational Advances in Artificial Intelligence}, 2019, pp. 750--757.

\bibitem{FactorizationMachine}
S.~Rendle, ``Factorization machines,'' in \emph{Proceedings of the 10th {IEEE} International Conference on Data Mining}, 2010, pp. 995--1000.

\bibitem{LSTM}
X.~Shi, Z.~Chen, H.~Wang, D.~Yeung, W.~Wong, and W.~Woo, ``Convolutional {LSTM} network: {A} machine learning approach for precipitation nowcasting,'' in \emph{Proceedings of the 29th Annual Conference on Neural Information Processing Systems}, 2015, pp. 802--810.

\bibitem{StableKT}
J.~Zhu, X.~Ma, and C.~Huang, ``Stable knowledge tracing using causal inference,'' \emph{IEEE Transactions on Learning Technologies}, vol.~17, pp. 124--134, 2024.

\bibitem{DCD}
F.~Wang, Z.~Huang, Q.~Liu, E.~Chen, Y.~Yin, J.~Ma, and S.~Wang, ``Dynamic cognitive diagnosis: An educational priors-enhanced deep knowledge tracing perspective,'' \emph{IEEE Transactions on Learning Technologies}, vol.~16, no.~3, pp. 306--323, 2023.

\bibitem{MRTKT}
J.~Cui, Z.~Chen, A.~Zhou, J.~Wang, and W.~Zhang, ``Fine-grained interaction modeling with multi-relational transformer for knowledge tracing,'' \emph{ACM Transactions on Information Systems}, vol.~41, no.~4, pp. 104:1--104:26, 2023.

\bibitem{OPKT}
L.~Qiu, M.~Zhu, and J.~Zhou, ``{OPKT}: Enhancing knowledge tracing with optimized pretraining mechanisms in intelligent tutoring,'' \emph{IEEE Transactions on Learning Technologies}, vol.~17, pp. 841--855, 2024.

\bibitem{GPT1}
A.~Radford, K.~Narasimhan, T.~Salimans, and I.~Sutskever, ``Improving language understanding by generative pre-training,'' \url{https://www.mikecaptain.com/resources/pdf/GPT-1.pdf}, 2018.

\bibitem{GPT2}
A.~Radford, J.~Wu, R.~Child, D.~Luan, D.~Amodei, I.~Sutskever \emph{et~al.}, ``Language models are unsupervised multitask learners,'' \emph{OpenAI Blog}, vol.~1, no.~8, p.~9, 2019.

\bibitem{GPT4}
J.~Achiam, S.~Adler, S.~Agarwal, L.~Ahmad, I.~Akkaya, F.~L. Aleman, D.~Almeida, J.~Altenschmidt, S.~Altman, S.~Anadkat \emph{et~al.}, ``{GPT-4} technical report,'' \emph{arXiv preprint arXiv:2303.08774}, 2023.

\bibitem{KQN}
J.~Lee and D.~Yeung, ``Knowledge query network for knowledge tracing: How knowledge interacts with skills,'' in \emph{Proceedings of the 9th International Conference on Learning Analytics and Knowledge}, 2019, pp. 491--500.

\bibitem{CL4KT}
W.~Lee, J.~Chun, Y.~Lee, K.~Park, and S.~Park, ``Contrastive learning for knowledge tracing,'' in \emph{Proceedings of the ACM Web Conference 2022}, 2022, pp. 2330--2338.

\bibitem{LBKT}
B.~Xu, Z.~Huang, J.~Liu, S.~Shen, Q.~Liu, E.~Chen, J.~Wu, and S.~Wang, ``Learning behavior-oriented knowledge tracing,'' in \emph{Proceedings of the 29th ACM SIGKDD Conference on Knowledge Discovery and Data Mining}, 2023, pp. 2789--2800.

\bibitem{qDKT}
S.~Sonkar, A.~S. Lan, A.~E. Waters, P.~Grimaldi, and R.~G. Baraniuk, ``{qDKT}: Question-centric deep knowledge tracing,'' in \emph{Proceedings of the 13th International Conference on Educational Data Mining}, 2020.

\bibitem{SemanticKT}
W.~Wang, H.~Ma, Y.~Zhao, F.~Yang, and L.~Chang, ``{SEEP}: Semantic-enhanced question embeddings pre-training for improving knowledge tracing,'' \emph{Information Sciences}, vol. 614, pp. 153--169, 2022.

\bibitem{ednet}
Y.~Choi, Y.~Lee, D.~Shin, J.~Cho, S.~Park, S.~Lee, J.~Baek, C.~Bae, B.~Kim, and J.~Heo, ``{EdNet}: {A} large-scale hierarchical dataset in education,'' in \emph{Proceedings of the 21st International Conference on Artificial Intelligence in Education}, 2020, pp. 69--73.

\bibitem{assistment2009}
L.~Razzaq, J.~Patvarczki, S.~F. Almeida, M.~Vartak, M.~Feng, N.~T. Heffernan, and K.~R. Koedinger, ``The assistment builder: Supporting the life cycle of tutoring system content creation,'' \emph{IEEE Transactions on Learning Technologies}, vol.~2, no.~2, pp. 157--166, 2009.

\bibitem{junyi}
H.~Chang, H.~Hsu, and K.~Chen, ``Modeling exercise relationships in e-learning: {A} unified approach,'' in \emph{Proceedings of the 8th International Conference on Educational Data Mining}, 2015, pp. 532--535.

\bibitem{CoKT}
T.~Long, J.~Qin, J.~Shen, W.~Zhang, W.~Xia, R.~Tang, X.~He, and Y.~Yu, ``Improving knowledge tracing with collaborative information,'' in \emph{Proceedins of the 15th {ACM} International Conference on Web Search and Data Mining}, 2022, pp. 599--607.

\bibitem{GPT-J}
B.~Wang, ``{Mesh-Transformer-JAX: Model-Parallel Implementation of Transformer Language Model with JAX},'' \url{https://github.com/kingoflolz/mesh-transformer-jax}, May 2021.

\bibitem{GPT3}
T.~B. Brown, B.~Mann, N.~Ryder, M.~Subbiah, J.~Kaplan, P.~Dhariwal, A.~Neelakantan, P.~Shyam, G.~Sastry, A.~Askell, S.~Agarwal, A.~Herbert{-}Voss, G.~Krueger, T.~Henighan, R.~Child, A.~Ramesh, D.~M. Ziegler, J.~Wu, C.~Winter, C.~Hesse, M.~Chen, E.~Sigler, M.~Litwin, S.~Gray, B.~Chess, J.~Clark, C.~Berner, S.~McCandlish, A.~Radford, I.~Sutskever, and D.~Amodei, ``Language models are few-shot learners,'' in \emph{Proceedings of the 34th Annual Conference on Neural Information Processing Systems}, 2020.

\bibitem{RoPE}
J.~Su, M.~H.~M. Ahmed, Y.~Lu, S.~Pan, W.~Bo, and Y.~Liu, ``Roformer: Enhanced transformer with rotary position embedding,'' \emph{Neurocomputing}, vol. 568, p. 127063, 2024.

\bibitem{pykt}
Z.~Liu, Q.~Liu, J.~Chen, S.~Huang, J.~Tang, and W.~Luo, ``{pyKT}: {A} {Python} library to benchmark deep learning based knowledge tracing models,'' in \emph{Proceedings of the 36th Annual Conference on Neural Information Processing Systems}, 2022.

\bibitem{EKT}
Q.~Liu, Z.~Huang, Y.~Yin, E.~Chen, H.~Xiong, Y.~Su, and G.~Hu, ``{EKT}: Exercise-aware knowledge tracing for student performance prediction,'' \emph{IEEE Transactions on Knowledge and Data Engineering}, vol.~33, no.~1, pp. 100--115, 2021.

\bibitem{UMAP}
L.~McInnes, J.~Healy, N.~Saul, and L.~Gro{\ss}berger, ``{UMAP}: Uniform manifold approximation and projection,'' \emph{Journal of Open Source Software}, vol.~3, no.~29, p. 861, 2018.

\end{thebibliography}

\begin{IEEEbiography}[{\includegraphics[width=1in,height=1.25in,clip,keepaspectratio]{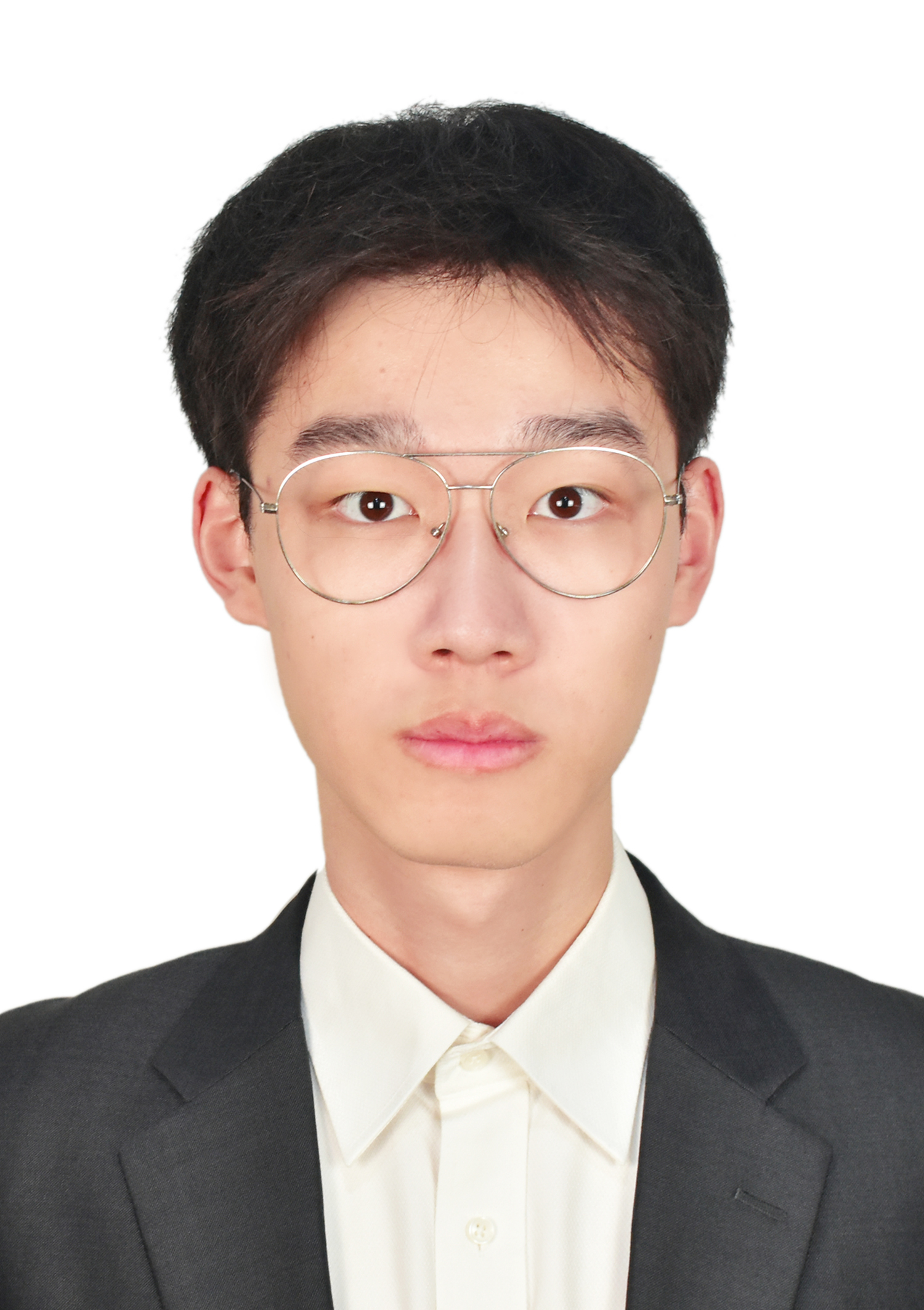}}]{Hao~Zhou} is a BSc student in the School of Computer Science and Engineering, Beihang University. His research interests include natural language processing and machine learning.
\end{IEEEbiography}
\begin{IEEEbiography}[{\includegraphics[width=1in,height=1.25in,clip,keepaspectratio]{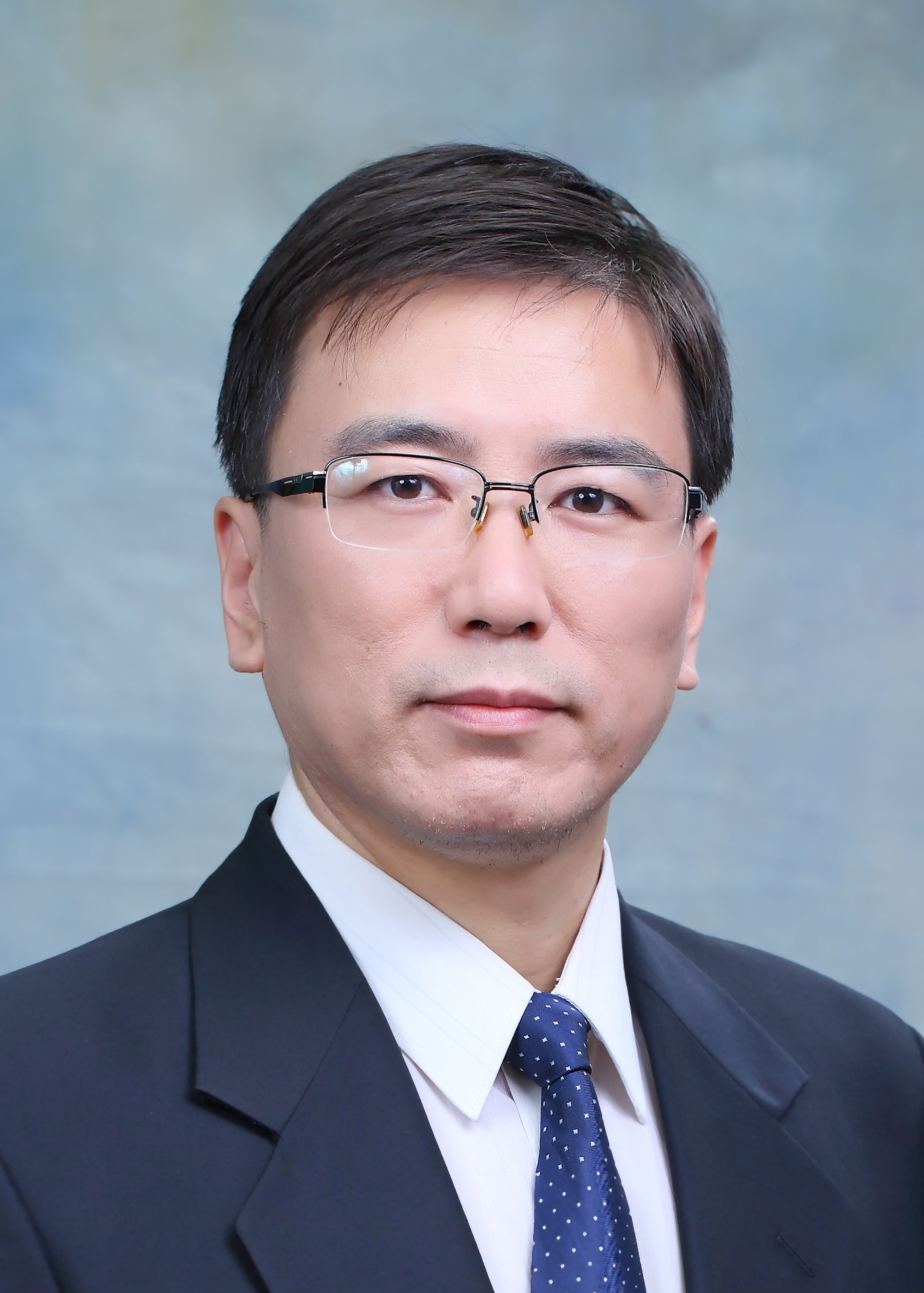}}]{Wenge~Rong}
	is a professor in the School of Computer Science and Engineering, Beihang University, China. He received his PhD from the University of Reading, UK, in 2010; MSc from Queen Mary College, University of London, UK, in 2003; and BSc from Nanjing University of Science and Technology, China, in 1996. He has many years of experience as a senior software engineer in numerous research projects and commercial software products. His area of research covers machine learning, natural language processing, and information management.
\end{IEEEbiography}
\begin{IEEEbiography}[{\includegraphics[width=1in,height=1.25in,clip,keepaspectratio]{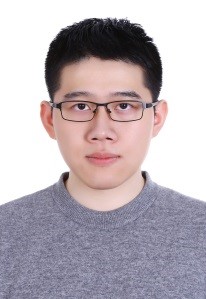}}]{Jianfei~Zhang} received the BSc degree from the
School of Computer Science of Engineering, Beihang University, in 2019. He is currently working toward a PhD degree with the School of Computer Science
and Engineering, Beihang University. His research interests include natural language processing and machine learning.
\end{IEEEbiography}
\begin{IEEEbiography}[{\includegraphics[width=1in,height=1.25in,clip,keepaspectratio]{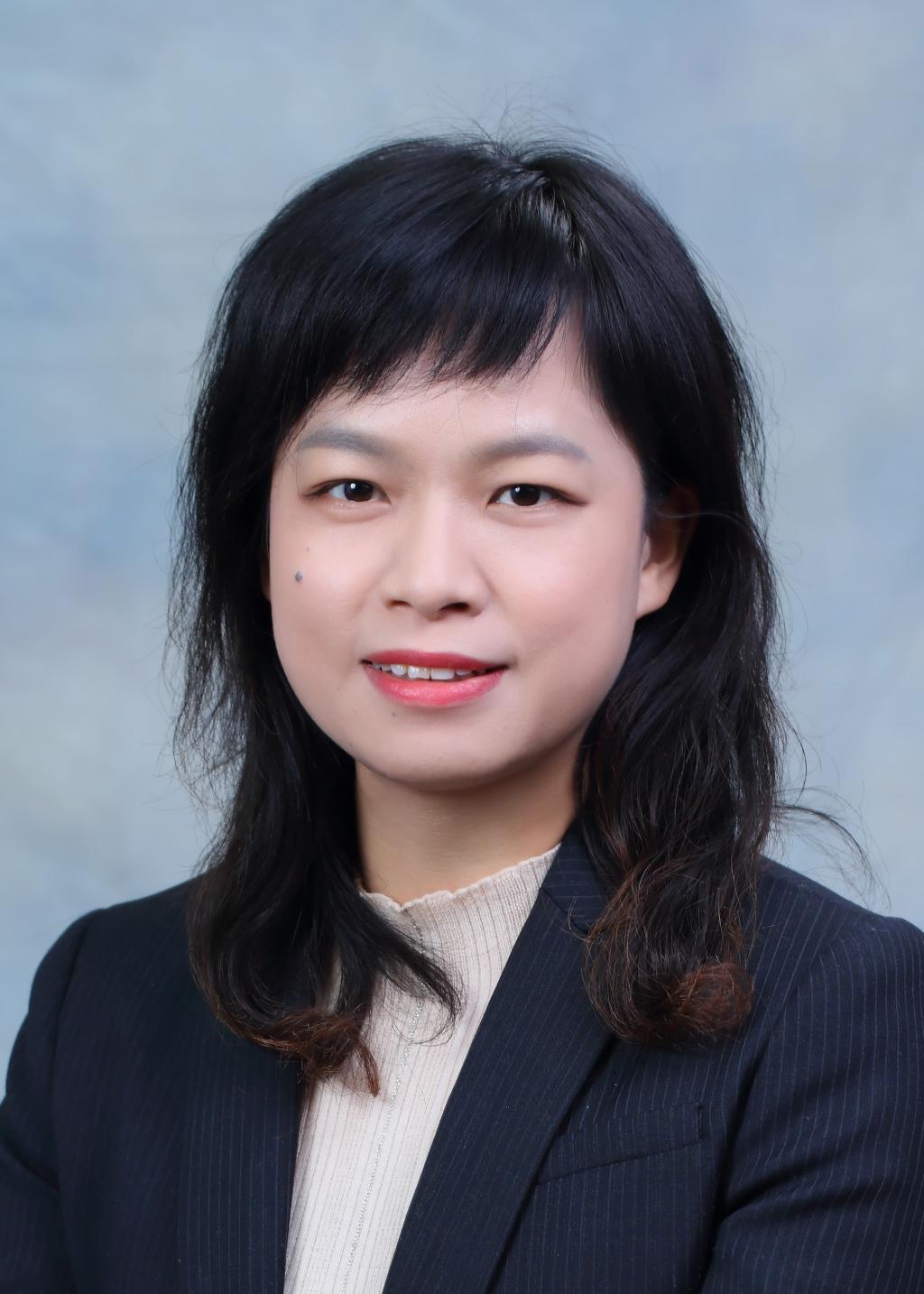}}]{Qing~Sun} is an associate professor in the School of Computer Science and Engineering, Beihang University. She received her Ph.D. from the Department of Information System of Beihang University in 2017; MSc and BSc from the School of Information Science of Beijing Normal University in 2008 and 2005, respectively. Her research interests include smart education, data mining, and knowledge engineering.
\end{IEEEbiography}

% \vspace{-6pt}
\begin{IEEEbiography}[{\includegraphics[width=1in,height=1.25in,clip,keepaspectratio]{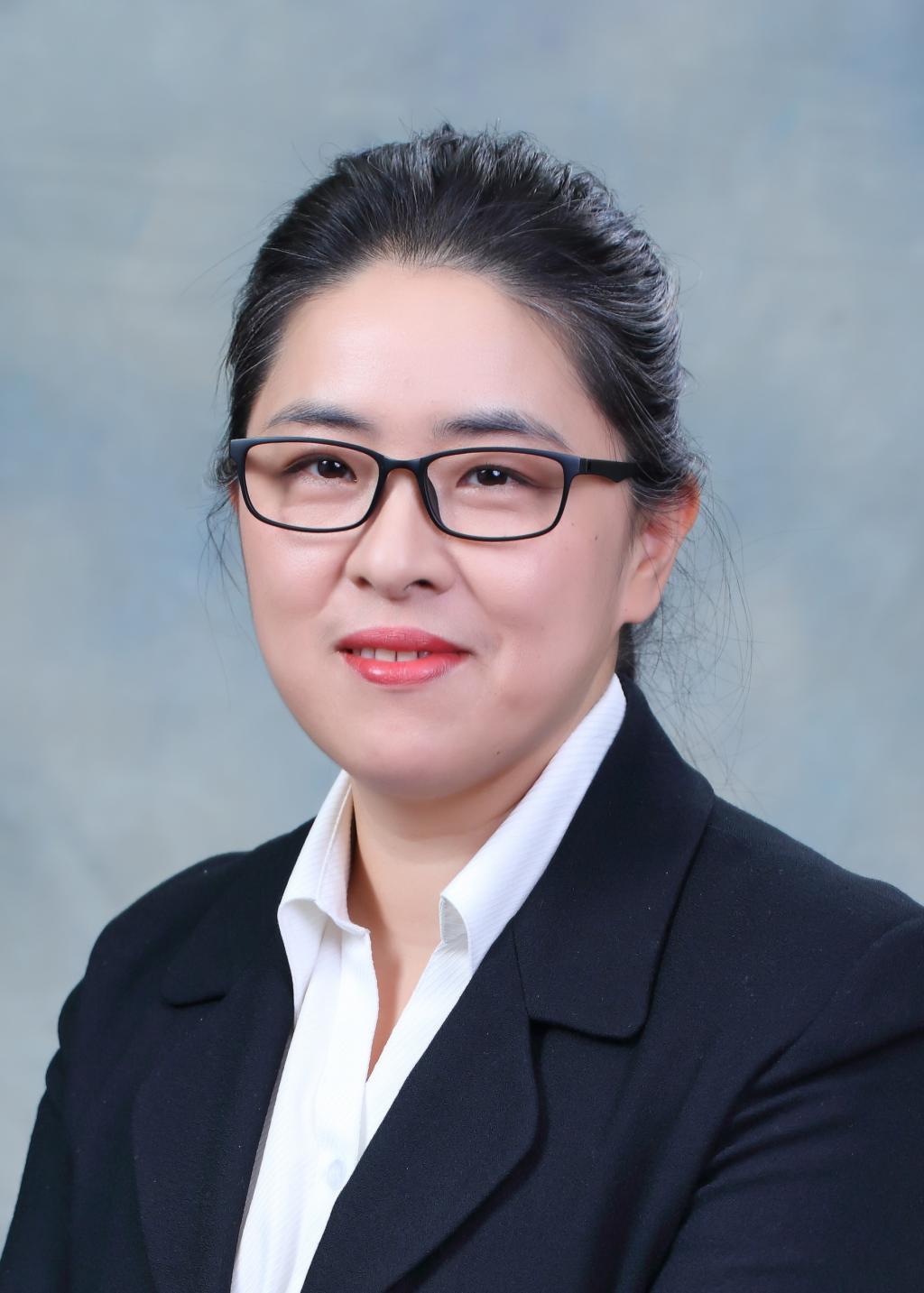}}]{Yuanxin~Ouyang}
	received her BSc and PhD degrees from Beihang University, in 1997 and 2005, respectively. She is a professor at School of Computer Science and Engineering, Beihang University, China. Her area of research covers recommender systems, data mining, and social networks.
\end{IEEEbiography}
% \vspace{-6pt}

\begin{IEEEbiography}[{\includegraphics[width=1in,height=1.25in,clip,keepaspectratio]{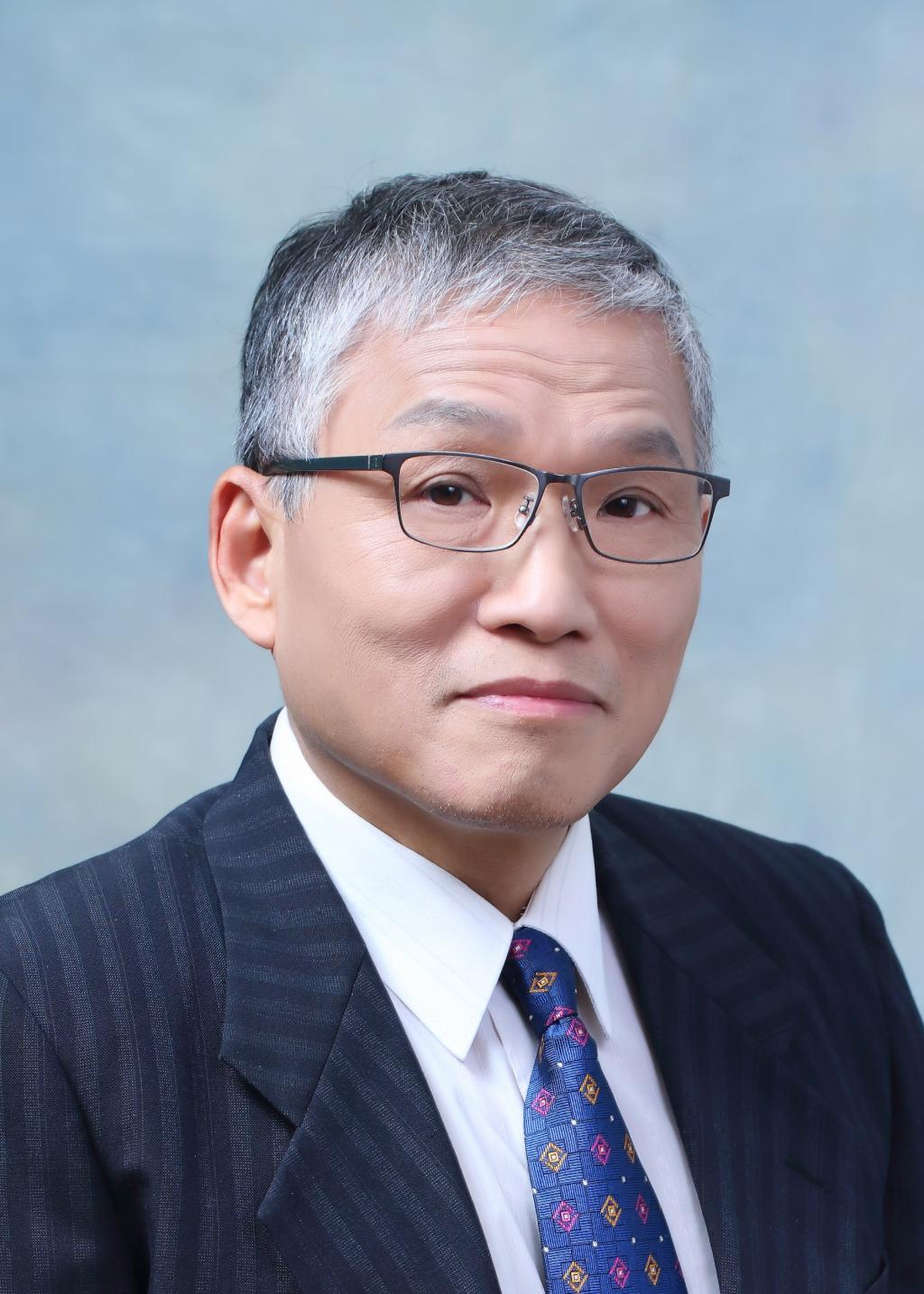}}]{Zhang~Xiong}
	is a professor with the School of Information Technology \& Management, University of International Business and Economics, and director of the Advanced Computer Application Research Engineering Center, National Educational Ministry of China. He has published more than 250 referred papers in international journals and conference proceedings and won a National Science and Technology Progress Award. His research interests span from smart cities, knowledge management, and information systems.
\end{IEEEbiography}

\end{document}